\documentclass{article}

\usepackage[preprint,nonatbib]{style}

\usepackage[ruled]{algorithm2e} % For algorithms

\usepackage{amsmath}
\usepackage{amssymb}
\usepackage{mathtools}
\usepackage{amsthm}
\usepackage{hyperref}
\usepackage{url}
\usepackage{stfloats}
\usepackage{enumitem}
\usepackage{wrapfig}
\usepackage{float}
\usepackage{placeins}
\usepackage{booktabs}
\usepackage{multirow}
\usepackage{colortbl}
\usepackage{caption}

\usepackage{media9}
\usepackage{siunitx}
\usepackage{pifont}
\usepackage{lipsum}
\usepackage{amsmath}
\usepackage{tcolorbox}
\usepackage{ragged2e}

\newcommand\mypara[1]{\vspace{1pt}\noindent\textbf{#1.}}

\definecolor{bestcell}{RGB}{255,205,205}
\definecolor{secondcell}{RGB}{255,229,191}
\newcommand{\best}[1]{\cellcolor{bestcell}\textbf{#1}}
\newcommand{\second}[1]{\cellcolor{secondcell}#1}

\title{Look-Before-Move: Narrative-Grounded World Visual Attention in Dynamic 3D Story Worlds}

\newcommand{\corrauthor}{\textsuperscript{\textdagger}}

\author{%
  \textbf{Jiaming Bian\textsuperscript{1} \quad Bingliang Li\textsuperscript{2} \quad Yuehao Wu\textsuperscript{3} \quad Pichao Wang\textsuperscript{4}\textsuperscript{*}} \\
  \textbf{Zhi Wang\textsuperscript{5} \quad Hailan Ma\textsuperscript{6} \quad Huadong Mo\textsuperscript{3}\corrauthor \quad Zhenhong Sun\textsuperscript{7}\corrauthor} \\
  \textsuperscript{1}Central South University\ \textsuperscript{2}Dependent Researcher\ \textsuperscript{3}University of New South Wales\ \textsuperscript{4}NVIDIA \\
  \textsuperscript{5}Nanjing University \quad \textsuperscript{6}Nanyang Technological University \quad \textsuperscript{7}Australian National University \\
  \texttt{bianjiaming@csu.edu.cn} \quad \texttt{\detokenize{{yuehao.wu, huadong.mo}@unsw.edu.au}} \\
  \texttt{lbingl@outlook.com} \quad \texttt{pichaowang@gmail.com} \quad \texttt{zhiwang@nju.edu.cn} \\
  \texttt{hailanma0413@gmail.com} \quad \texttt{zhenhong.sun@anu.edu.au}
}

\begin{document}

\maketitle

\begingroup
\renewcommand{\thefootnote}{\null}
\footnotetext{\textsuperscript{*}This work is not related to the author’s position at NVIDIA.\quad\corrauthor~Co-corresponding authors}
\endgroup

\begin{abstract}
As embodied AI and world models increasingly operate in dynamic 3D environments, visual perception must move beyond passively interpreting given observations toward actively deciding what to observe. 
We study this problem through camera planning in dynamic 3D story worlds, where the camera must not only generate smooth motion, but also decide what visual evidence should be acquired before it moves. 
We formulate this capability as \textbf{Narrative-Grounded World Visual Attention}, where the camera acts as an embodied observer that determines what to observe, how to compose the observation, and how to shift attention over time under narrative intent and physical 3D constraints. 
To realize this capability, we propose \textbf{Look-Before-Move}, a camera planning framework that separates observation specification from motion execution. 
It first builds a Semantic Observation Contract to convert directorial intent into executable visual constraints, then performs Monte Carlo Viewpoint Search to find narrative-compliant and geometrically feasible viewpoints, and finally applies Semantic Trajectory Grounding to connect selected viewpoints into continuous, collision-aware, and temporally coherent camera motion. 
We further construct a dynamic 3D Story World Benchmark based on \textit{StoryBlender}, covering 50 stories, 457 scenes, and 1585 shots with animated characters, semantic scene configurations, and executable 3D environments. 
Experiments show that our framework improves subject perception, intent consistency, and trajectory quality over representative baselines, demonstrating the importance of organizing visual attention before generating camera motion.
Project page and code are available at: \url{https://engineeringai-lab.github.io/Look-Before-Move/}.
\end{abstract}

\section{Introduction}

As embodied AI~\cite{hu2024scenecraft,yang2026sceneweaver,wang2026harnesspai} and world models~\cite{liu2025worldcraft,lin2026pat3d} increasingly operate inside executable 3D environments, visual perception can no longer be treated as a passive input stream. 
Unlike conventional image or video understanding, where visual evidence is already given, an agent in a native 3D world must actively determine what to observe before planning how to move. 
This connects to autonomous camera control~\cite{lino2015intuitive} and camera trajectory planning~\cite{dehghanian2026camera}, but shifts the focus from smooth trajectory generation to observation-driven camera planning.
This question is especially important in dynamic 3D story worlds, where recent systems have begun to construct editable narrative environments with controllable layouts, assets, character actions, and spatial-temporal dynamics~\cite{rao2023dynamic,huang2024story3d,li2026storyblender}. 
Here, a 3D story world is an executable, editable narrative environment with geometry, identified characters and objects, spatial layouts, scripted actions, and time-varying states in shared 3D coordinates.
A camera in such worlds is not simply a geometric sensor that maximizes visibility or avoids collisions. 
It acts as the eye of the world: selecting narrative subjects, emphasizing actions, revealing spatial relations, and shifting attention as the story unfolds. 
For example, when a story describes an assassin pointing a pistol, the camera should not merely capture any person; it should identify the acting subject, preserve the action cue, and compose the scene according to its narrative meaning.
We define this capability as \textbf{Narrative-Grounded World Visual Attention}: the camera organizes what to observe, how to observe it, and when to transition attention within a physically constrained 3D world.

Realizing this capability requires solving three coupled problems. 
First, high-level directorial intent must be converted into visual requirements, such as target subjects, semantic relations, visibility conditions, composition preferences, and action cues. 
Second, the system must search for viewpoints that satisfy both narrative relevance and physical feasibility, since a semantically appropriate view may be invalid due to occlusion, collision, poor scale, or infeasible camera placement. 
Third, selected viewpoints must be grounded into camera motion that tracks dynamic actors, avoids obstacles, and preserves temporal coherence. 
These requirements go beyond camera-controllable video synthesis~\cite{wang2024motionctrl,he2024cameractrl,yang2024direct} and text-to-camera trajectory generation~\cite{zhang2025gendop,li2024director3d,courant2024exceptional}, where the objective is to follow a camera motion, synthesize a trajectory, or maintain visual consistency. 
Together, they advance camera planning toward the semantic-spatial-temporal organization of visual evidence.

Existing camera planning methods provide foundations, but they often assume that the observation objective is already specified. 
Rule-based and geometry-driven methods encode cinematographic heuristics, visibility constraints, collision avoidance, and smooth path generation~\cite{dehghanian2026camera,lino2015intuitive}. 
Recent learning-based~\cite{bahmani2025vd3d,xu2024cavia} and generative methods~\cite{wang2024motionctrl,he2024cameractrl,yang2024direct,xu2024camco} model camera motion priors or synthesize camera-controllable videos by injecting trajectory representations into video generation models. 
Several works analyze camera motion and 3D camera control in videos~\cite{zhao2022particlesfm,bahmani2025ac3d,lin2026towards}, while recent 3D cinematography systems explore auto-regressive trajectory generation~\cite{zhang2025gendop}, diffusion-based camera and scene generation~\cite{li2024director3d}, text-to-camera trajectory generation~\cite{courant2024exceptional}, and multi-agent film automation~\cite{xu2025filmagent}. 
These methods improve feasible trajectory generation, camera-controllable synthesis, or autonomous cinematography, but they do not address how an intent should be inferred from narrative intent and grounded in a physically executable 3D world.

To address this gap, we propose \textbf{Look-Before-Move}, a narrative-grounded camera planning framework built on a simple principle: before a camera can move intelligently, it must first know what visual evidence it is supposed to acquire. 
Instead of directly generating a trajectory from a story instruction, our framework first constructs a Semantic Observation Contract that converts directorial intent into executable visual constraints. 
It then performs Monte Carlo Viewpoint Search to identify viewpoints that are both narrative-compliant and geometrically feasible, followed by Semantic Trajectory Grounding to connect selected viewpoints into continuous camera motion. 
This decomposition turns narrative-grounded world visual attention into three executable steps: specifying the intended observation, selecting feasible viewpoints, and grounding them into motion.

A systematic study of this problem also requires an executable evaluation environment. 
Although recent 3D story-world generation methods provide editable narrative environments, real-world filming is costly and difficult to control. Additionally, 2D video datasets or diffusion-generated videos cannot reliably determine whether a camera plan is physically feasible, whether a better viewpoint exists, or whether a failure results from semantic misunderstanding, geometric occlusion, or motion instability. 
We therefore construct a dynamic 3D Story World Benchmark based on \textit{StoryBlender}~\cite{li2026storyblender}. 
The benchmark covers 50 stories, 457 scenes, and 1585 shots, and supports controlled evaluation along three dimensions: subject perception, intent consistency, and trajectory quality.
Our experiments show that Look-Before-Move improves over representative camera trajectory planning baselines across all three evaluation dimensions. 
The results support our central claim: in dynamic 3D story worlds, effective camera planning should begin with observation before motion generation. 
% By first grounding narrative intent into explicit visual constraints, then searching for feasible viewpoints, and finally generating executable trajectories, our framework provides a practical step toward controllable visual attention in embodied 3D worlds.

In summary, our main contributions are as follows:
\begin{itemize}[leftmargin=*, noitemsep, nolistsep]
    \item[$\bullet$] We formulate \textbf{Narrative-Grounded World Visual Attention} as a camera planning capability in dynamic 3D story worlds, reframing the problem from passively interpreting given observations to intent-driven visual evidence acquisition.
    \item[$\bullet$] We propose \textbf{Look-Before-Move}, a narrative-grounded camera planning framework that decomposes camera planning into observation specification, viewpoint search, and trajectory grounding.

    \item[$\bullet$] We construct a dynamic 3D Story World Benchmark on executable 3D scenes, supporting controlled evaluation of subject perception, intent consistency, and trajectory quality.
\end{itemize}

\section{Related Work}

\mypara{Camera Trajectory Planning}
Autonomous camera control has long studied generating feasible and visually pleasing viewpoints or trajectories in virtual environments. 
Early methods rely on cinematography rules, geometric constraints, visibility reasoning, collision avoidance, and smooth paths~\cite{dehghanian2026camera,lino2015intuitive}. 
Recent generative methods extend camera control to video generation by injecting trajectory representations, such as extrinsic matrices or Plucker coordinates, into diffusion models~\cite{wang2024motionctrl,he2024cameractrl,yang2024direct}, or by improving camera-controllable synthesis through multi-view and 3D-aware consistency~\cite{xu2024camco,xu2024cavia,bahmani2025vd3d}. 
Generation work covers entity guidance~\cite{sun2024eggen}, diversity~\cite{liu2026beyond}, and video alignment~\cite{wang2024lift}.
However, many approaches~\cite{zhao2022particlesfm,bahmani2025ac3d,lin2026towards} operate primarily in pixel space and lack access to executable 3D embodied worlds. 
Recent 3D cinematography methods explore auto-regressive trajectory generation~\cite{zhang2025gendop}, diffusion-based camera and scene generation~\cite{li2024director3d}, and multi-agent film automation~\cite{xu2025filmagent}. 
Different from these works, our goal is not only to synthesize a trajectory, but to ground the intended observation from narrative intent before camera motion.

\mypara{3D Story Worlds}
The construction of 3D story worlds provides the foundation for controllable narrative visualization. 
Engine-based methods show that virtual environments can support dynamic storyboard generation and pre-visualization~\cite{rao2023dynamic}. 
Recent LLM-based systems use code generation or structured reasoning to synthesize open-domain 3D scenes with editable layouts and assets~\cite{hu2024scenecraft,yang2026sceneweaver,liu2025worldcraft,lin2026pat3d}. 
Sketch-guided generation~\cite{xu2024sketch2scene} also supports interactive 3D scenes.
Other works introduce character action scripting and narrative behavior modeling, enabling 3D environments to express story events instead of static layouts~\cite{huang2024story3d}. 
\textit{StoryBlender} further maintains cross-scene asset consistency and produces editable dynamic 3D storyboards with spatial-temporal dynamics~\cite{li2026storyblender}. 
These methods make it possible to construct narrative 3D worlds, but remain open how such worlds should be observed. 
Our work complements 3D story-world by introducing a camera planning framework that organizes visual attention inside executable story worlds.

\mypara{LLM Reflection and Agentic Planning}
Large language models have been used as agents that reason, reflect, and refine outputs through feedback. 
Frameworks such as Reflexion and Self-Refine show that language models can evaluate and revise their own generations~\cite{shinn2023reflexion,madaan2023self}. 
Works extend reflection with dynamic instructions, visual-language reasoning, experience accumulation, tree search, and multi-agent collaboration~\cite{liu2025instruct,cheng2025vision,zhao2024expel,zhou2023language,guo2025mirror}. 
These approaches are effective in text-centric or interactive environments, including narrative settings~\cite{wu2025towards}. 
Related work explores instruction-guided medical VQA~\cite{wu2024detecting}, feature-space knowledge enhancement~\cite{wu2026injection}, hierarchical decision-making~\cite{hu2025divide}, and multi-round medical reasoning~\cite{wu2026marag}.
In dynamic 3D worlds, however, reflection must be grounded in physical constraints: a proposed view may be occluded, the camera may collide with the scene, the subject may leave the frame, or the trajectory may become unstable. 
Embodied reasoning and visual reflection studies suggest that planning feedback should be tied to visual and geometric evidence~\cite{lin2025navcot,jian2025look,wei2025perception}. 
In our framework, reflection is used as physical verification for viewpoint selection and trajectory grounding rather than as a verbal memory mechanism.

\section{Method}

\mypara{Problem Statement}
Let $\mathcal{W}=(\mathcal{G},\mathcal{A},\mathcal{L},\mathcal{M})$ denote an executable dynamic 3D story world, where $\mathcal{G}$ is the scene geometry, $\mathcal{A}$ contains characters, objects, and semantic attributes, $\mathcal{L}$ describes the spatial layout, and $\mathcal{M}$ records animations and object motions. Given a directorial intent $\mathcal{I}$ and a temporally ordered event sequence $\mathcal{S}=(s_1,\ldots,s_N)$, scene-to-camera trajectory planning aims to produce a continuous camera trajectory for each event:
\begin{equation}
    \tau_i = \pi(\mathcal{W}, \mathcal{I}, \mathcal{S}_{\leq i}),
\end{equation}
where $\tau_i:[0,1]\rightarrow \mathrm{SE}(3)$ is an executable camera motion and $\pi(\cdot)$ is the planning policy. Here, $\mathrm{SE}(3)$ represents the mathematical space of 3D positions and orientations. Executing $\tau_i$ in the 3D world produces an observation clip:
\begin{equation}
    c_i=\mathcal{R}(\mathcal{W},\tau_i,s_i),
\end{equation}
and the final output is an ordered observation sequence $\mathcal{E}=(c_1,\ldots,c_N)$.

Unlike static viewpoint prediction or low-level trajectory generation, this task aims to organize \textit{world visual attention} by deciding what should be observed, emphasized, and revealed over time. It requires solving two coupled problems: \textit{spatial viewpoint composition}, which selects intent-compliant and visible viewpoints under geometric constraints, and \textit{temporal motion coordination}, which aligns these viewpoints with feasible camera motion relative to dynamic subjects and narratives.

To address this task, we propose \textbf{Look-Before-Move}, a collaborative multi-agent camera planning framework consisting of three components. First, \textbf{3D Story World Benchmark} provides executable dynamic scenes with geometry, semantic entities, animations, shot-level configurations, and rendering feedback for controlled camera planning evaluation. Second, \textbf{Look} addresses \textit{spatial viewpoint composition} by establishing an Observation Contract to search for safe, visible, and intent compliant viewpoints before any camera motion is committed. Third, \textbf{Move} addresses \textit{temporal motion coordination} through grounding selected viewpoints into continuous camera trajectories that follow dynamic subjects, avoid collisions, and seamlessly align with narrative progression. 
% Together, these components enable \textbf{Look-Before-Move} to turn high-level directorial intent into controllable and physically feasible world visual attention in dynamic 3D story worlds.
\begin{figure*}[!t]
  \centering
  \includegraphics[width=\linewidth]{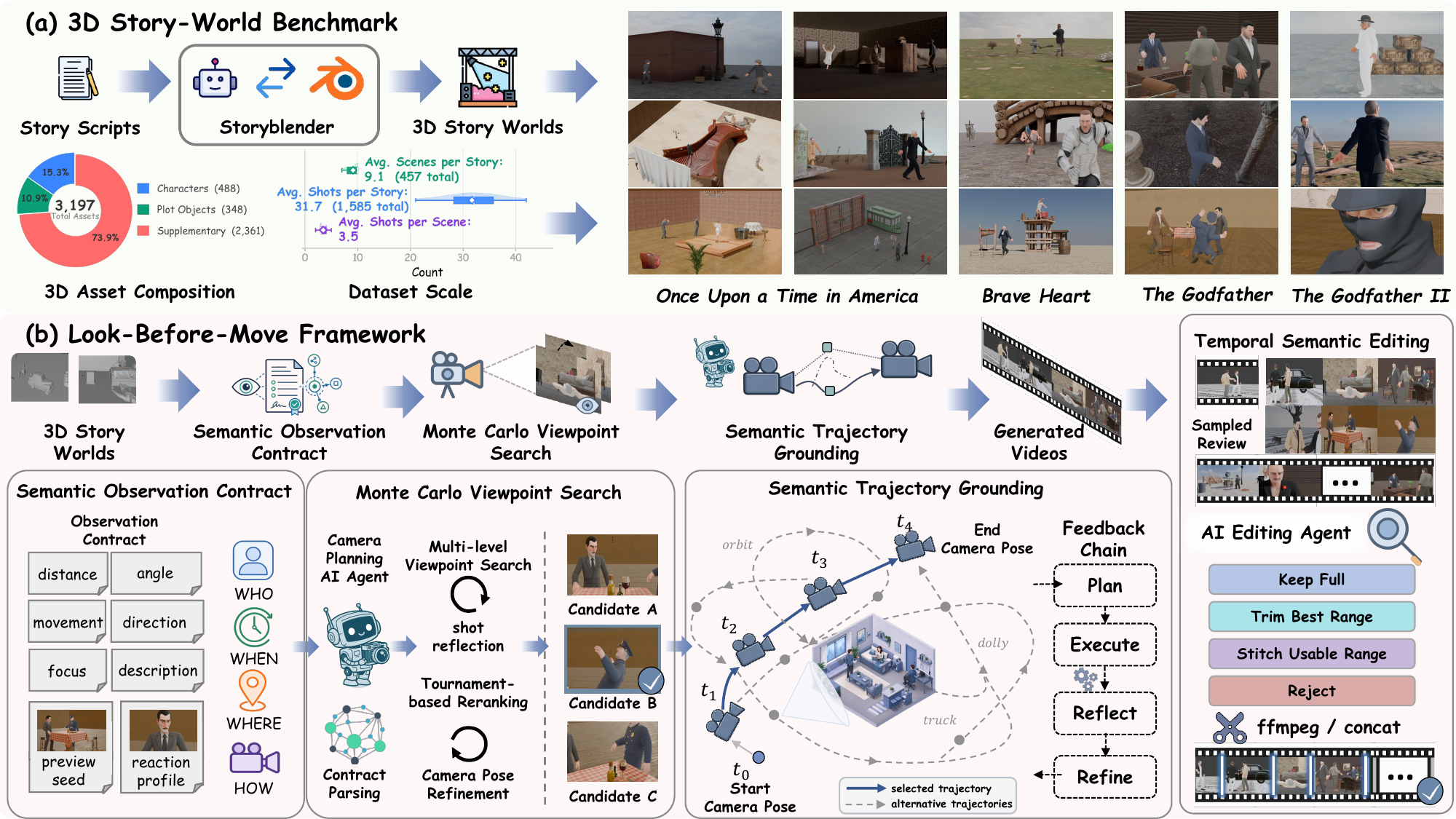}
  \caption{Overview of the benchmark construction and Look-Before-Move framework. 
(a) We build a dynamic 3D story-world benchmark using \textit{StoryBlender}, covering narrative scripts, controllable scenes, character animations, camera annotations, and dataset statistics. 
(b) Look-Before-Move converts narrative intent into an observation contract, searches and refines feasible viewpoints in the executable 3D world, and grounds selected views into temporally coherent camera trajectories.}
  \label{fig:story}
\end{figure*}

\subsection{3D Story World Benchmark}
\label{sec: Benchmark}
To study narrative-grounded camera planning, we require a benchmark where story intent, dynamic events, character identities, and physical 3D geometry are jointly controllable and verifiable.

\mypara{3D World Building} Existing 2D video datasets or diffusion-generated scenes often suffer from spatial hallucinations, identity drift, and a lack of physical grounding, severely limiting their utility for rigorous camera planning. To overcome these limitations, we construct our executable benchmark using \textit{StoryBlender}~\cite{li2026storyblender}, an engine-verified 3D generation framework that guaranties strict \textit{inter-shot consistency} and \textit{explicit editability}. Unlike 2D pixel-space representations, our virtual world relies on a persistent, physically grounded simulation. Formally, this world is represented as $\mathcal{W}=(\mathcal{G},\mathcal{A},\mathcal{L},\mathcal{M})$, corresponding to scene geometry, canonical character assets, spatial layout, and dynamic action sequences. By materializing entities in a unified 3D coordinate space, the environment provides the camera with precise geometric boundaries and semantic perception capabilities. Building upon this foundation, we integrate these engine-verified scenes, animations, and temporal events to formalize our dataset as $\mathcal{D}=\{(\mathcal{W}_j,\mathcal{I}_j,\mathcal{S}_j, \mathcal{Y}^{\mathrm{scene}}_j)\}_{j=1}^{M}$, where $\mathcal{I}_j$ denotes the directorial intent, $\mathcal{S}_j$ represents the event progression, and $\mathcal{Y}^{\mathrm{scene}}_j$ stores the comprehensive metadata necessary for programmatic evaluation. This establishes a reliable mapping between narrative intent and physical rendering, ensuring that camera decisions receive accurate visual and geometric feedback.

\mypara{Benchmark} 
Addressing the lack of dynamic evaluation, we propose an Executable 3D World Benchmark for world visual attention. 
Spanning 50 stories, the dataset comprises 457 dynamic scenes and 1585 shots. 
To support narrative environments, it includes 3197 3D assets (836 plot-related assets and 2361 supplementary assets for scene decoration), featuring an average of 9.76 characters per story and 1.44 character actions per shot. 
In the evaluation, given the 3D world $\mathcal{W}$, intent $\mathcal{I}$, and event process $\mathcal{S}=(s_1,\ldots,s_N)$, the model must output a continuous camera trajectory $\tau_i:[0,1]\rightarrow \operatorname{SE}(3)$ for each unit $s_i$. 
Because our dataset is built on 3D environments rather than static pose predictions, it allows models to make spatial-temporal decisions and undergo closed-loop engine verification. 
We conduct quantitative evaluations across three dimensions: (1) \textbf{Subject Perception}, assessing the ability to capture targets and avoid physical occlusions; (2) \textbf{Intent Consistency}, verifying whether viewpoints adhere to cinematic semantics; and (3) \textbf{Trajectory Quality}, examining motion smoothness, spatial-temporal dynamics, and execution feasibility.

\subsection{Look for Spatial Viewpoint Composition}
Before the camera can move, it must first determine what the story requires it to see. 
The \textit{Look} stage addresses this problem by converting high-level directorial intent into an adaptive observation contract that specifies the subjects, actions, spatial relations, visibility requirements, and composition preferences to be satisfied by candidate viewpoints. 
To this end, a group of specialized perception and planning agents collaborate over narrative instructions and rendered scene evidence, establishing a semantic-geometric perception interface between the story world and the camera.

\mypara{Semantic Observation Contract} 
In complex 3D dynamic environments, the system first needs an accurate understanding of the current physical and visual state. To achieve this, we introduce the Perception Agent. By interacting with the 3D world, this agent samples and renders multi-view preview images around the target character to obtain intuitive environmental feedback. These multi-view previews not only capture the character's pose and spatial position but also reveal potential environmental occlusions, thus extracting scene state representations rich in visual and geometric information from the observation unit $s_i$. This intelligent perception mechanism based on real rendering establishes a solid foundation for subsequent semantic intent parsing and planning.

Faced with abstract high-level directorial intent, the camera struggles to execute it directly, necessitating a precise perception-based mapping mechanism. To this end, the Planning Agent, upon receiving the perception results, proposes an Observation Contract. Through semantic parsing $\Phi$, this agent converts the abstract intent $\mathcal{I}$ and the scene state of the current observation unit $s_i$ into a concrete observation contract $\mathcal{O}_i$:
\begin{equation}
\mathcal{O}_i = \{a^*, C_{\text{spatial}}, T_{\text{action}}\} = \Phi(\mathcal{I}, s_i)
\end{equation}
where $a^*$ denotes the observable target subject, $C_{\text{spatial}}$ represents spatial composition requirements, and $T_{\text{action}}$ is the temporal action constraint. This structured contract not only specifies ``what to look at" and ``how to look at it" for the camera but also effectively narrows the state space for subsequent viewpoint search, successfully establishing a mapping between semantic intent and 3D geometry to provide prior guidance for precise camera control.

\mypara{Monte Carlo Viewpoint Search} 
Merely establishing observation targets is insufficient to guaranty the usability of the framing; the camera must also identify the optimal visual landing point in the executable 3D world $\mathcal{W}$. To achieve this, we design the Camera Reflection mechanism, a self-evaluation process based on multi-agent collaboration, viewpoint search, and fine-tuning. First, based on the observation contract $\mathcal{O}_i$, the system drives multiple Camera Agents to perform Monte Carlo-based Multi-level Viewpoint Search, generating a large number of candidate shot boards in the large space. Subsequently, the Evaluation Agent executes Tournament-based Reranking on these candidate viewpoint sets $\mathcal{V}$, progressively eliminating low-quality shots through competitive scoring to select the winning initial optimal viewpoint $v_{\text{init}}^*$:
\begin{equation}
v_{\text{init}}^* = \arg\max_{v \in \mathcal{V}} \left( \lambda_1 \text{Vis}(v, a^*) + \lambda_2 \text{Comp}(v, C_{\text{spatial}}) - \lambda_3 \text{Occ}(v, \mathcal{W}) \right)
\end{equation}
where $\text{Vis}(\cdot)$ evaluates subject visibility, $\text{Comp}(\cdot)$ measures composition rationality, and $\text{Occ}(\cdot)$ calculates physical occlusion caused by the environment. 

Following this competitive selection, the Reflection Agent refines the winning viewpoint through localized camera-pose adjustment. 
Starting from $v_{\text{init}}^*$, it applies small perturbations to the camera position, orientation, and focal parameters, and re-renders the scene to check whether the target subject remains visible, the action cue is preserved, and the composition contract is satisfied. 
Refinements that introduce collision, severe occlusion, excessive subject truncation, or unstable framing are rejected. 
Through this render-evaluate-adjust loop, the system corrects minor visual imperfections, such as off-center subjects, poor scale, partial occlusion, or weak spatial relation expression, and obtains the final viewpoint $v_i^*$.

In summary, the \textit{Look} stage converts narrative intent into a physically executable spatial observation decision. 
It first builds a semantic observation contract from narrative intent and rendered scene evidence, then searches and refines candidate viewpoints through Monte Carlo sampling, tournament reranking, and render-based reflection. 
The resulting viewpoint $v_i^*$ grounds ``what should be observed'' in the executable 3D world and serves as the visual anchor for subsequent trajectory grounding in the \textit{Move} stage.

\subsection{Move for Temporal Motion Coordination }
Once the observation contract and initial optimal viewpoint are determined, the system transitions from the ``searching" for shots to the ``executing" motion stage of camera trajectory planning. At this stage, the planning and editing agents work collaboratively to generate continuous, stable, and executable motion trajectories in the 3D scene, a process we term \textbf{Semantic Trajectory Grounding}.
The whole process of Semantic Trajectory Grounding can be divided into Trajectory Planning, Trajectory Reflection, and Temporal Semantic Editing.

\mypara{Trajectory Planning} 
After acquiring a sequence of high-quality discrete shots (keyframes) $\{v_t^*\}$, generating smooth, continuous trajectories that adapt to dynamic subject movements is the core of achieving high-quality visual presentation. The Planning Agent uses the discrete shots as key nodes to solve for the optimal continuous trajectory $\tau_i^*$ in the joint spatial-temporal space:
\begin{equation}
\tau_i^* = \arg\min_{\tau} \sum_{t} \left( \|\tau(t) - v_t^*\|^2 + \gamma \text{Cost}_{\text{smooth}}(\dot{\tau}(t), \ddot{\tau}(t)) \right)
\end{equation}
This objective function aims to minimize the deviation between the continuous trajectory $\tau(t)$ and the discrete key shots $v_t^*$, while strictly constraining the camera's kinematic properties (such as velocity $\dot{\tau}(t)$ and acceleration $\ddot{\tau}(t)$) through the penalty term $\text{Cost}_{\text{smooth}}$. Through this relay strategy of first generating discrete shots by the Camera Agent, then generating continuous trajectories by the Planning Agent, the system can plan camera trajectories that both stably track the target and exhibit excellent smoothness.

\mypara{Trajectory Reflection} 
Although the initially planned trajectories are theoretically feasible, they may still encounter unexpected physical obstacles or intent deviations in complex dynamic scenes, necessitating strict closed-loop verification. To address this, the Evaluation Agent intervenes again to trigger the Trajectory Reflection mechanism. It simulates execution and comprehensively scores $R(\tau_i)$ the entire generated trajectory $\tau_i$ in a real 3D rendering environment:
\begin{equation}
R(\tau_i) = \alpha P_{\text{subj}}(\tau_i) + \beta I_{\text{intent}}(\tau_i, \mathcal{I}) + \delta Q_{\text{traj}}(\tau_i)
\end{equation}
The formula comprehensively evaluates the trajectory from three dimensions: subject perception $P_{\text{subj}}$, intent consistency $I_{\text{intent}}$, and motion quality $Q_{\text{traj}}$. Only trajectory segments with scores exceeding a threshold are retained. This agent-driven closed-loop verification mechanism actively identifies and corrects sub-optimal or unexecutable actions, greatly enhancing the robustness of camera movements and ensuring the absolute reliability of trajectories in complex environments.

\mypara{Temporal Semantic Editing} When dealing with complex scenes containing multiple event streams, a single viewpoint is often insufficient to fully convey all information, making reasonable shot scheduling particularly important. At this stage, the Editing Agent takes over and schedules all observation clip sets $\{c_i\}$ in the temporal dimension. It screens and splices them to generate the final observation sequence $\mathcal{E}^*$:
\begin{equation}
\mathcal{E}^* = \arg\max_{\mathcal{E} \subset \{c_i\}} \left( \sum_{c_i \in \mathcal{E}} U(c_i) - \eta \sum_{j=1}^{|\mathcal{E}|-1} D_{\text{trans}}(c_j, c_{j+1}) \right)
\end{equation}
where $U(c_i)$ measures the information transmission utility and semantic concentration of a single clip, and $D_{\text{trans}}(\cdot)$ calculates the visual incoherence penalty for adjacent viewpoint switches. By balancing clip quality and switching smoothness, the Editing Agent condenses discrete observation clips into a high-quality visual sequence, thereby conveying high-level directorial intent as accurately and vividly as possible.

In summary, the \textit{Move} stage transforms selected viewpoints into temporally coordinated camera behavior. 
It first connects discrete observation anchors into smooth continuous trajectories, then verifies their physical feasibility and narrative consistency through closed-loop trajectory reflection. 
Finally, temporal semantic editing schedules verified clips into a coherent observation sequence. 
As a result, the system converts ``how to observe'' into executable camera motion that remains stable, feasible, and aligned with the evolving story.

\section{Experiment}
\label{sec:experiment}
\subsection{Implementation Details}
\mypara{Benchmark Metrics}
We evaluate camera planning from three complementary dimensions: \textbf{Subject Perception (SP)}, \textbf{Intent Consistency (IC)}, and \textbf{Trajectory Quality (TQ)}. 
SP focuses on whether the generated camera behavior can correctly capture the intended subjects under dynamic scene conditions. 
It measures target visibility, framing completeness, and robustness to physical occlusions. 
IC evaluates whether the resulting views are semantically aligned with the given filming intent, including shot scale, narrative focus, action relevance, and consistency with the event description. 
TQ assesses the execution quality of the generated camera motion, including trajectory smoothness, motion stability, transition coherence, and temporal continuity. 
All metrics are normalized to a 0--100 scale, where higher scores indicate better performance. 
Detailed metric definitions and computation protocols are provided in the \textbf{Appendix \ref{Appendix Benchmark Details}}.

\mypara{Setup and Evaluator Models}
Our rendering and physics validation environment is built on Blender, which provides executable 3D scenes, camera control, collision checking, and frame-level rendering for evaluation. 
We use Gemini-3-Flash-Preview~\cite{google2025gemini3flash} for agent reasoning, reflection, and instruction generation. 
For subject perception evaluation, we use YOLO11x~\cite{yolo11_ultralytics} together with ORB~\cite{rublee2011orb} to assess target detection, visibility, and frame-level consistency. 
For intent consistency, we use Qwen2.5-VL-7B-Instruct~\cite{Qwen2.5-VL} to perform blind text-video alignment scoring between the rendered camera video and the corresponding filming instruction. 
To ensure a fair comparison, all methods are evaluated using the same story segments, prompts, renderer, evaluator models, and segment-count weighting protocol. 
All videos are rendered at 24 FPS on a workstation equipped with one NVIDIA RTX 5090 GPU. Details are provided in the \textbf{Appendix \ref{Appendix Metric computation}} .

\mypara{Datasets}
We evaluate all methods on the cinematic test set introduced in this work, which covers indoor and outdoor scenes, multi-character interactions, object-centric events, and temporally evolving story segments. 
Each scene is paired with natural language filming instructions specifying the intended subject, narrative focus, action cue, and shot-level visual requirements, enabling zero-shot evaluation of text-to-3D camera scheduling. 
This setting tests whether a method can translate high-level directorial intent into executable camera behavior that decides both what to observe and how to organize motion over time; details are provided in \textbf{Appendix \ref{Appendix Data hierarchy}}.

% Keep experimental pages at natural spacing, including the gap below Figure 2.
\suppressfloats[t]

\begin{table*}[t]
\centering
\footnotesize
\caption{Main quantitative comparison with segment-count weighting. Scores include subject perception, intent consistency, trajectory quality, user study, and overall; higher is better. \colorbox{bestcell}{first best} and \colorbox{secondcell}{second best} mark the strongest two entries.}
\label{tab:main_results}
\setlength{\tabcolsep}{3pt}
\renewcommand{\arraystretch}{1.28}
\resizebox{\linewidth}{!}{%
\begin{tabular}{cl|cccc|cccc|cccc|c|c}
\hline
\multirow{2}{*}{Set} & \multirow{2}{*}{Methods} 
& \multicolumn{4}{|c|}{Subject perception} 
& \multicolumn{4}{c|}{Intent consistency} 
& \multicolumn{4}{c|}{Trajectory quality} 
& \multirow{2}{*}{\begin{tabular}{@{}c@{\;}c@{}}\begin{tabular}{@{}c@{}}User\\Study\end{tabular} & \raisebox{0.45ex}{$\uparrow$}\end{tabular}}
& \multirow{2}{*}{Overall $\uparrow$} \\
\cline{3-14}
& & SP1 $\uparrow$ & SP2 $\uparrow$ & SP3 $\uparrow$ & Mean $\uparrow$ 
& IC1 $\uparrow$ & IC2 $\uparrow$ & IC3 $\uparrow$ & Mean $\uparrow$ 
& TQ1 $\uparrow$ & TQ2 $\uparrow$ & TQ3 $\uparrow$ & Mean $\uparrow$ & & \\
\hline
\multirow{4}{*}{\rotatebox[origin=c]{90}{\textbf{Baseline}}} 
& CCD 
& \second{88.45} & \second{66.72} & \second{69.87} & \second{75.01} 
& 58.41 & \second{85.71} & \best{24.86} & 55.96 
& 40.01 & \second{86.05} & 72.88 & \second{66.15} & \second{3.21} & \second{65.62} \\

& GenDoP+RGBD 
& 82.74 & 57.80 & 67.27 & 69.27 
& 74.30 & 76.52 & 22.08 & 57.30 
& 32.64 & 80.99 & 73.29 & 62.00 & 2.92 & 62.75 \\

& GenDoP+noRGBD & 83.75 & 59.14 & 67.00 & 69.96 & 72.53 
& 78.17 & \second{24.17} & \second{57.91} & 36.13 & 81.25 
& \best{73.94} & 63.41 & 3.01 & 63.63 \\

& FilmAgent-adapted
& 75.47 & 55.24 & 65.92 & 65.54
& 78.18 & 59.90 & 24.77 & 54.28
& 40.08 & 73.82 & 63.04 & 59.09 & --- & 59.64 \\
\hline
\multirow{3}{*}{\rotatebox[origin=c]{90}{\textbf{Director}}} 
& ET+Director A 
& 59.69 & 36.82 & 59.72 & 52.08 
& \best{85.92} & 63.17 & 10.80 & 52.92 
& \second{40.81} & 56.96 & 67.64 & 55.00 & 2.28 & 53.28 \\

& ET+Director B 
& 61.34 & 38.60 & 60.43 & 53.46 
& \second{85.90} & 62.98 & 17.20 & 54.98 
& 39.69 & 61.11 & 67.18 & 55.87 & 2.35 & 54.71 \\

& ET+Director C 
& 57.40 & 43.54 & 51.22 & 50.72 
& 79.20 & 69.80 & 19.63 & 55.00 
& 38.86 & 57.51 & 67.59 & 54.55 & 2.25 & 53.26 \\
\hline
\multirow{1}{*}[1ex]{\rotatebox[origin=c]{90}{\scriptsize\textbf{Ours}}}
& \rule[-1.1ex]{0pt}{4.2ex}Look-Before-Move 
& \best{90.02} & \best{66.97} & \best{72.59} & \best{76.53} 
& 78.15 & \best{90.30} & 23.26 & \best{63.90} 
& \best{62.44} & \best{87.73} & \second{73.92} & \best{74.65} & \best{4.08} & \best{71.70} \\
\hline
\end{tabular}
}
\end{table*}

\begin{figure}[t]
\centering
\includegraphics[width=\linewidth]{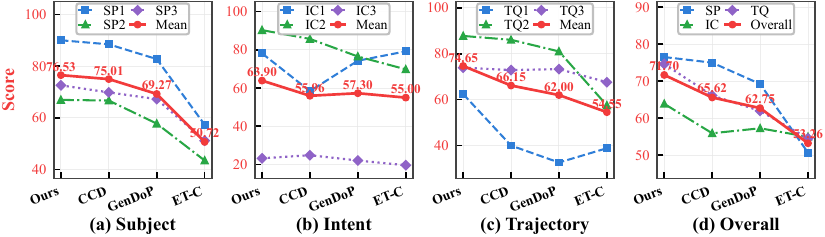}
\caption{Fine-grained quantitative comparison across subject perception, intent consistency, trajectory quality, and overall performance. Look-Before-Move improves the segment-weighted score while maintaining stronger motion-related metrics than the compared camera planning baselines. The panel exposes where each method succeeds or fails beyond the single aggregate score.}
\label{fig:main_quantitative_panel}
\end{figure}

\subsection{Main Results}

Based on the benchmark protocol above, we compare Look-Before-Move with representative camera planning and trajectory generation baselines, including Cinematographic Camera Diffusion \cite{jiang2024cinematographic}, GenDoP \cite{zhang2025gendop}, and Exceptional Trajectories \cite{courant2024exceptional}. We report results from two complementary perspectives: namely quantitative comparison and qualitative visual analysis.
FilmAgent-adapted preserves FilmAgent's~\cite{xu2025filmagent} cinematographer--director negotiation, shot proposals, and iterative review, using a deterministic adapter to convert its observation plan into Blender-executable trajectories without manual selection or correction. Under the same stories, rendering pipeline, and evaluation protocol, Look-Before-Move achieves 71.70 Overall versus 59.64 for FilmAgent-adapted, with gains of 10.99 SP, 9.62 IC, and 15.56 TQ points.

\begin{figure}[t]
\centering
\noindent
\begin{minipage}[h]{0.40\linewidth}
\centering
\includegraphics[width=0.95\linewidth]{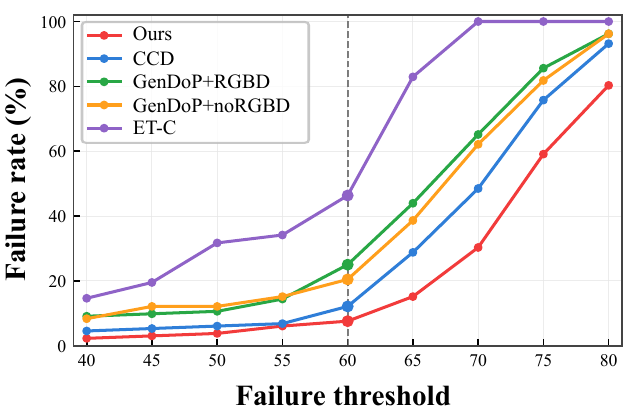}
\captionof{figure}{Failure-case counts under unified evaluation. Lower curves indicate fewer segments below each threshold.}
% , showing stronger robustness across strict and relaxed settings.}
\label{fig:failure_case_count_comparison}
\end{minipage}
\hfill
\begin{minipage}[h]{0.58\linewidth}
\centering
\captionof{table}{Ablation results with segment-count weighting between different variants. Each row removes one component under the same evaluator, showing its impact on SP, IC, TQ, and overall score.}
\label{tab:ablation}
\scriptsize
\resizebox{\linewidth}{!}{%
\begin{tabular}{lcccc}
\hline
Variant & SP $\uparrow$ & IC $\uparrow$ & TQ $\uparrow$ & Overall $\uparrow$ \\
\hline
Ours & \textbf{76.53} & \textbf{63.90} & \textbf{74.65} & \textbf{71.70} \\
w/o MLS (Fast) & 67.54 & 53.28 & 60.91 & 60.52 \\
w/o VLM-R & 75.30 & 60.82 & 71.86 & 69.11 \\
w/o TG & 69.05 & 53.24 & 60.19 & 60.74 \\
w/o SHA & 68.28 & 53.57 & 60.23 & 60.54 \\
w/o PCSJ & 64.24 & 55.00 & 59.90 & 59.45 \\
\hline
\end{tabular}
}
\end{minipage}
\end{figure}

\mypara{Quantitative Evaluation}
Table~\ref{tab:main_results} and Figures~\ref{fig:main_quantitative_panel} and~\ref{fig:failure_case_count_comparison} show that Look-Before-Move achieves the best overall performance under the unified segment-weighted evaluation protocol. 
Our method obtains an Overall score of 71.70, outperforming the strongest complete baseline CCD by 6.08 points, with clear gains in both intent consistency and trajectory quality.
These results show that dynamic 3D camera planning requires more than keeping subjects visible or generating smooth motion. 
Existing baselines often address only one aspect of the problem, such as subject perception, shot-size matching, or trajectory generation, but struggle to jointly satisfy narrative intent, physical feasibility, and temporal coherence. 
By first identifying narrative-compliant viewpoints and then grounding them into executable trajectories, Look-Before-Move provides a more complete solution to world visual attention. 
The results support our central argument: camera planning should be formulated as observation-driven visual evidence organization rather than direct trajectory generation.

\begin{figure}[t]
\centering
\begin{minipage}[t]{0.56\linewidth}
\centering
\includegraphics[width=0.96\linewidth]{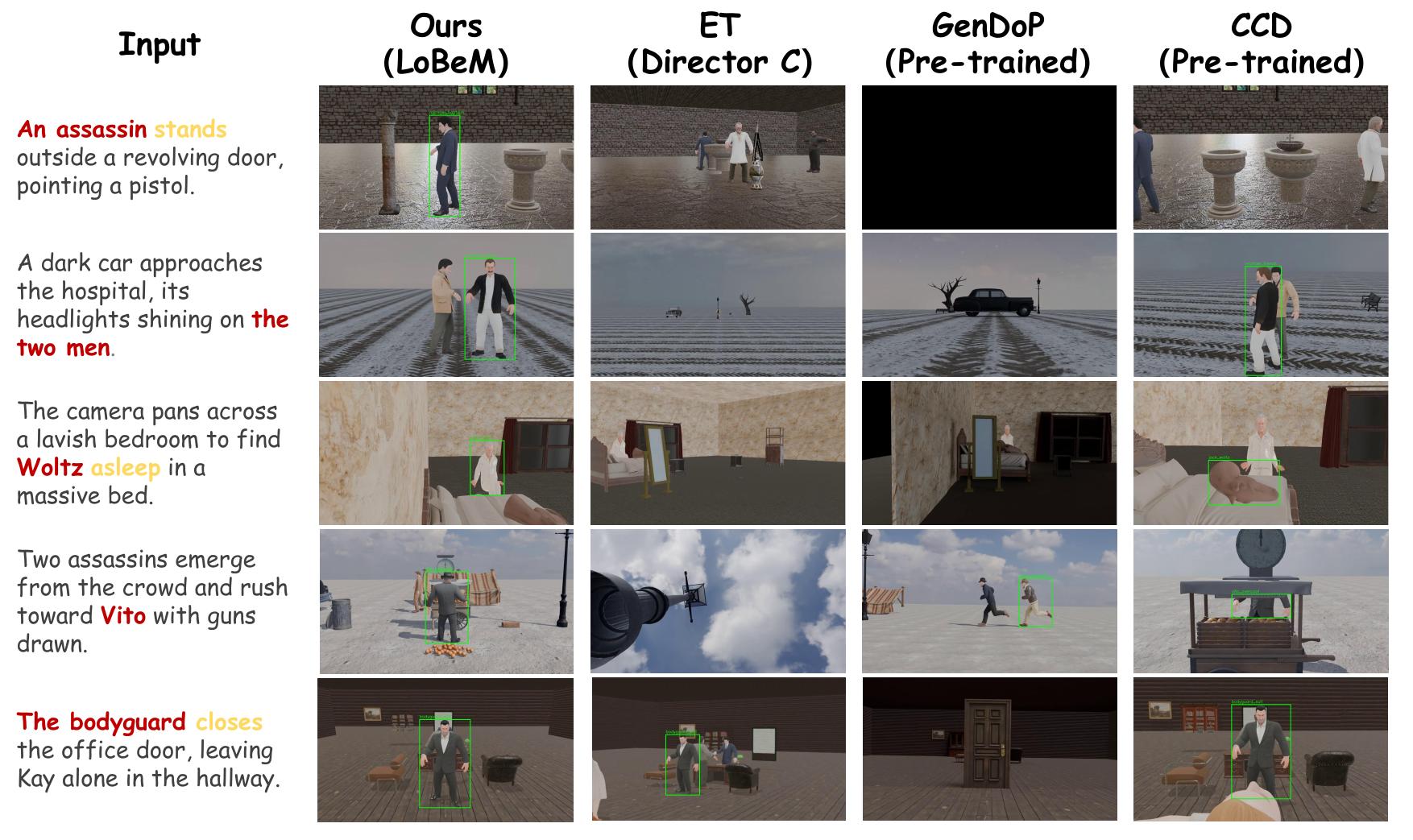}
\caption{Qualitative comparison on representative story segments. Rows pair narrative instructions with outputs from our method and baselines, testing subject identity, action cues, spatial context, and story-relevant framing.}
\label{fig:qualitative_comparison}
\end{minipage}
\hfill
\begin{minipage}[t]{0.42\linewidth}
\centering
\includegraphics[width=\linewidth]{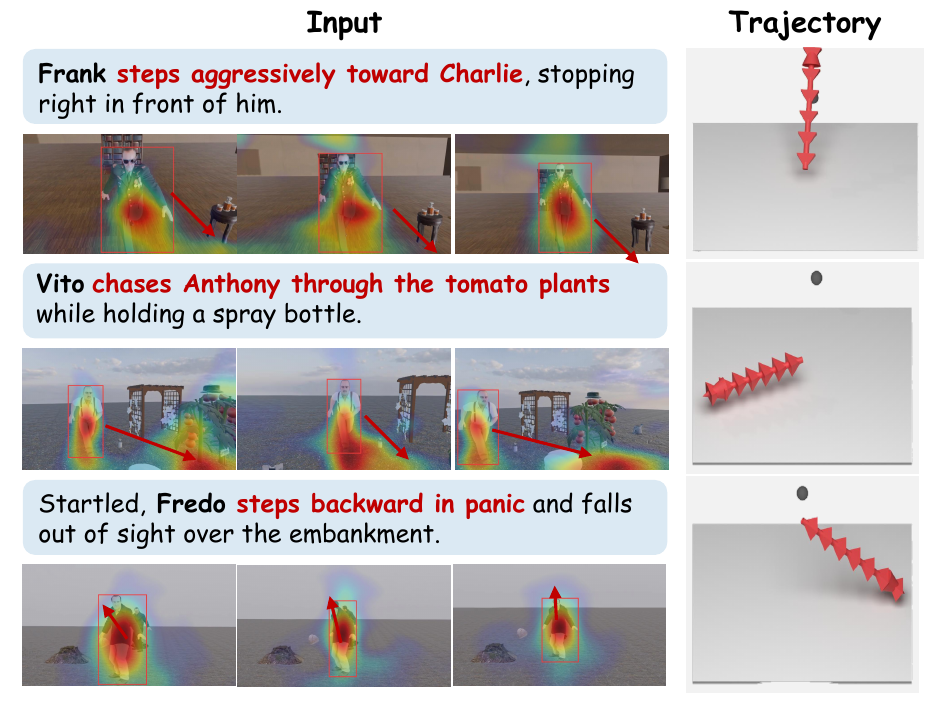}
\caption{Qualitative visualization of world visual attention. 
Attention maps and trajectories show narrative intent grounded in evidence before execution.}
\label{fig:attention_qualitative}
\end{minipage}
\end{figure}

\mypara{Qualitative Evaluation}
Figure~\ref{fig:qualitative_comparison} shows qualitative results across story segments. 
In the assassin case, Look-Before-Move identifies the intended character, renders the subject at close-up scale, and preserves the action cue. 
In the car and hospital scenes, it selects viewpoints that capture interactions while maintaining stable framing. 
In the bedroom wide shot, the generated view keeps both the bed and the sleeping character within the narrative focus, preparing the subsequent plot where the character awakens. 
These examples show that our method does not merely generate plausible frames, but selects views that are physically valid and narratively specific.
Figure~\ref{fig:attention_qualitative} further illustrates how the framework maintains narrative-grounded visual attention over time. 
When the main subject moves across shots, the camera continues to track the protagonist, preserve action cues, and adapt its viewpoint to the evolving scene context. 
This behavior is enabled by converting the instruction into an observation contract and validating candidate views in the executable 3D world before trajectory grounding. 
As a result, Look-Before-Move reduces viewpoint drift and translates textual intent into stable visual evidence before camera motion is executed.

\begin{figure}[t]
\centering
\includegraphics[width=\textwidth]{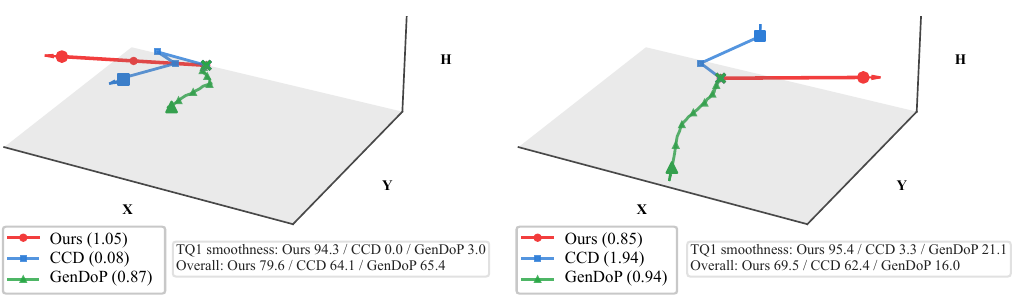}
\caption{Additional comparison of semantic trajectory quality. Look-Before-Move achieves stronger semantic alignment without sacrificing executable motion quality.}
\label{fig:appendix_semantic_trajectory}
\par
\includegraphics[width=\linewidth]{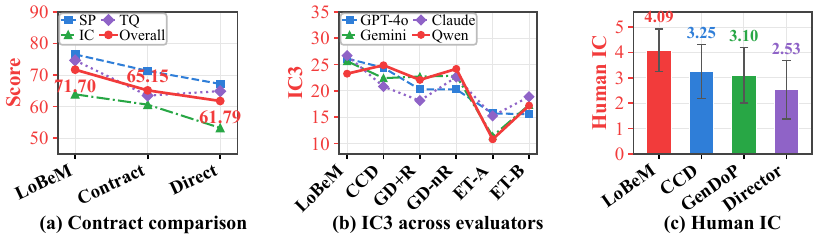}
\caption{Contract comparison (a), IC3 across evaluators (b), and human IC with SD error bars (c). Full results are in Appendices~\ref{app:cr_contract_budget}--\ref{app:cr_human_ic}.}
\label{fig:cr_contract_comparison}
\end{figure}

\subsection{Additional Results}

\mypara{Ablation Study}
Table~\ref{tab:ablation} evaluates five components under shared dataset, prompts, renderer, evaluator, and metrics. 
MLS, VLM-R, TG, SHA, and PCSJ denote Multi-level Monte Carlo Search, VLM Reflection, Trajectory Grounding, Semantic Height Adjust, and Pre-continuity Story Judge. 
The Fast variant is \textit{w/o MLS}, where Monte Carlo search and reranking are disabled.
The results show that viewpoint search and temporal grounding are the sources of improvement. 
Removing MLS causes the large drop, reducing the overall score from 71.70 to 60.52, confirming the importance of candidate expansion and reranking before motion planning. 
VLM-R improves intent consistency through visual-language feedback, while TG, SHA, and PCSJ are crucial for maintaining trajectory quality and temporal coherence. 
Overall, the ablation validates the core design of Look-Before-Move: narrative-compliant viewpoint selection must be coupled with executable motion grounding.

\mypara{Contract Analysis}
Figure~\ref{fig:cr_contract_comparison}(a) compares Direct Trajectory Generation (Direct), Observation Contract + Trajectory Generation (Contract), and LoBeM-full (LoBeM) under identical inputs, model settings, rendering, and evaluation; full results appear in Table~\ref{app:cr_contract_table} (Appendix~\ref{app:cr_contract}).
The first two use the same trajectory generator and differ only in whether the Observation Contract is used.
Compared with direct generation, the Observation Contract improves Overall by 3.36 points, SP by 4.13 points, and IC by 7.35 points, isolating the benefit of structuring what should be observed before motion planning.
LoBeM-full further improves Overall by 6.55 points, demonstrating the aggregate benefits of search, geometric verification, grounding, and reflection.

\mypara{Evaluation Validation}
Figure~\ref{fig:cr_contract_comparison}(b) places LoBeM first under the three independent IC3 evaluators and third under the original Qwen, with identical inputs and protocols (Appendix~\ref{app:cr_independent_ic3}).
Human IC also favors LoBeM (4.09), above CCD (3.25), GenDoP (3.10), and Director (2.53) (Figure~\ref{fig:cr_contract_comparison}(c); Appendix~\ref{app:cr_human_ic}).

\mypara{Trajectory Visualization}
Figure~\ref{fig:appendix_semantic_trajectory} provides an additional quantitative view of the trade-off between semantic quality and trajectory quality.
The comparison complements the main table by separating whether a method succeeds because it chooses story-relevant evidence, because it produces smooth executable motion, or because it balances both.
Look-Before-Move is designed to improve this balance: the Look stage improves semantic evidence acquisition, while the Move stage prevents high-quality still viewpoints from becoming unstable motion plans.

\mypara{Abstract Intent Recovery}
Using only executable Contract fields, abstract-intent recovery reaches 77.78\% for Abstract-A and 73.15\% for its paraphrase Abstract-B, compared with 16.67\% for uniform six-way prediction (Table~\ref{app:cr_abstract_recovery_table}).
These results show that abstract intents leave recoverable information in executable observation constraints after free-form intent text is removed, while concrete instructions are easier to recover (87.04\%).
Detailed controls and examples appear in Appendix~\ref{app:cr_abstract_intent}.

\mypara{Efficiency}
Separate end-to-end profiling on one RTX 3090 with Blender 4.5.5 LTS gives 26.50 seconds per effective video second for Look-Before-Move versus 18.16 for CCD, including rendering and evaluation (Table~\ref{app:cr_runtime_table}).
This is a $1.46\times$ runtime ratio for a 6.08-point Overall gain; detailed cost profiling appears in Appendix~\ref{app:cr_efficiency}.

\section{Conclusion}
We introduced narrative-grounded world visual attention as a novel paradigm for camera planning within dynamic 3D environments, moving beyond passive trajectory generation toward the active organization of visual evidence. 
To realize this capability, we proposed Look-Before-Move as a collaborative planning framework that effectively translates abstract directorial intent into executable camera behaviors by decoupling spatial viewpoint composition and temporal motion coordination. 
To facilitate systematic research, we constructed an executable 3D story-world benchmark that provides rigorous geometric and semantic feedback for closed-loop verification. 
Extensive experiments validate that our approach significantly outperforms existing methods across subject perception, intent consistency, and trajectory quality. 
Ultimately, this research bridges the critical gap between high-level cinematic intent and foundational physical execution, paving the way for fully autonomous and intelligent visual agents in complex virtual worlds, especially in long-horizon narrative generation and embodied world modeling. 
These results position camera control as active perception, requiring agents to reason jointly about narrative evidence, 3D observability, and executable motion.

\section*{Acknowledgements and Disclosures}
We received no funding or other third-party support for this work during the past 36 months. We declare no competing interests and no financial relationships outside the submitted work that could be perceived as influencing this research.

\bibliographystyle{ieeetr}
\bibliography{reference}

@inproceedings{li2026storyblender,
  title={Storyblender: Inter-shot consistent and editable 3d storyboard with spatial-temporal dynamics},
  author={Li, Bingliang and Sun, Zhenhong and Bian, Jiaming and Wu, Yuehao and Wang, Yifu and Li, Hongdong and Bian, Yatao and Mo, Huadong and Dong, Daoyi},
  booktitle={European Conference on Computer Vision},
  pages={130--147},
  year={2026},
  organization={Springer}
}

@article{shinn2023reflexion,
  title={Reflexion: Language agents with verbal reinforcement learning},
  author={Shinn, Noah and Cassano, Federico and Gopinath, Ashwin and Narasimhan, Karthik and Yao, Shunyu},
  journal={Advances in neural information processing systems},
  volume={36},
  pages={8634--8652},
  year={2023}
}

@article{madaan2023self,
  title={Self-refine: Iterative refinement with self-feedback},
  author={Madaan, Aman and Tandon, Niket and Gupta, Prakhar and Hallinan, Skyler and Gao, Luyu and Wiegreffe, Sarah and Alon, Uri and Dziri, Nouha and Prabhumoye, Shrimai and Yang, Yiming and others},
  journal={Advances in neural information processing systems},
  volume={36},
  pages={46534--46594},
  year={2023}
}

@inproceedings{liu2025instruct,
  title={Instruct-of-reflection: Enhancing large language models iterative reflection capabilities via dynamic-meta instruction},
  author={Liu, Liping and Zhang, Chunhong and Wu, Likang and Zhao, Chuang and Hu, Zheng and He, Ming and Fan, Jianping},
  booktitle={Proceedings of the 2025 Conference of the Nations of the Americas Chapter of the Association for Computational Linguistics: Human Language Technologies (Volume 1: Long Papers)},
  pages={9956--9978},
  year={2025}
}

@inproceedings{cheng2025vision,
  title={Vision-language models can self-improve reasoning via reflection},
  author={Cheng, Kanzhi and Li, Yantao and Xu, Fangzhi and Zhang, Jianbing and Zhou, Hao and Liu, Yang},
  booktitle={Proceedings of the 2025 Conference of the Nations of the Americas Chapter of the Association for Computational Linguistics: Human Language Technologies (Volume 1: Long Papers)},
  pages={8876--8892},
  year={2025}
}

@inproceedings{zhao2024expel,
  title={Expel: Llm agents are experiential learners},
  author={Zhao, Andrew and Huang, Daniel and Xu, Quentin and Lin, Matthieu and Liu, Yong-Jin and Huang, Gao},
  booktitle={Proceedings of the AAAI Conference on Artificial Intelligence},
  volume={38},
  number={17},
  pages={19632--19642},
  year={2024}
}

@article{zhou2023language,
  title={Language agent tree search unifies reasoning acting and planning in language models},
  author={Zhou, Andy and Yan, Kai and Shlapentokh-Rothman, Michal and Wang, Haohan and Wang, Yu-Xiong},
  journal={arXiv preprint arXiv:2310.04406},
  year={2023}
}

@article{guo2025mirror,
  title={Mirror: Multi-agent intra-and inter-reflection for optimized reasoning in tool learning},
  author={Guo, Zikang and Xu, Benfeng and Wang, Xiaorui and Mao, Zhendong},
  journal={arXiv preprint arXiv:2505.20670},
  year={2025}
}

@inproceedings{wu2025towards,
  title={Towards enhanced immersion and agency for llm-based interactive drama},
  author={Wu, Hongqiu and Wu, Weiqi and Xu, Tianyang and Zhang, Jiameng and Zhao, Hai},
  booktitle={Proceedings of the 63rd Annual Meeting of the Association for Computational Linguistics (Volume 1: Long Papers)},
  pages={11166--11182},
  year={2025}
}

@inproceedings{jian2025look,
  title = {Look Again, Think Slowly: Enhancing Visual Reflection in Vision-Language Models},
  author = {Jian, Pu and Wu, Junhong and Sun, Wei and Wang, Chen and Ren, Shuo and Zhang, Jiajun},
  editor = {Christodoulopoulos, Christos and Chakraborty, Tanmoy and Rose, Carolyn and Peng, Violet},
  booktitle = {Proceedings of the 2025 Conference on Empirical Methods in Natural Language Processing},
  month = nov,
  year = {2025},
  address = {Suzhou, China},
  publisher = {Association for Computational Linguistics},
  url = {https://aclanthology.org/2025.emnlp-main.470/},
  doi = {10.18653/v1/2025.emnlp-main.470},
  pages = {9251--9270},
  ISBN = {979-8-89176-332-6}
}

@article{wei2025perception,
  title={Perception in reflection},
  author={Wei, Yana and Zhao, Liang and Lin, Kangheng and Yu, En and Peng, Yuang and Dong, Runpei and Sun, Jianjian and Wei, Haoran and Ge, Zheng and Zhang, Xiangyu and others},
  journal={arXiv preprint arXiv:2504.07165},
  year={2025}
}

@incollection{rao2023dynamic,
  title={Dynamic storyboard generation in an engine-based virtual environment for video production},
  author={Rao, Anyi and Jiang, Xuekun and Guo, Yuwei and Xu, Linning and Yang, Lei and Jin, Libiao and Lin, Dahua and Dai, Bo},
  booktitle={ACM SIGGRAPH 2023 Posters},
  pages={1--2},
  year={2023}
}

@inproceedings{hu2024scenecraft,
  title={SceneCraft: An LLM Agent for Synthesizing 3D Scenes as Blender Code.},
  author={Hu, Ziniu and Iscen, Ahmet and Jain, Aashi and Kipf, Thomas and Yue, Yisong and Ross, David A and Schmid, Cordelia and Fathi, Alireza},
  booktitle={ICML},
  pages={19252--19282},
  year={2024}
}

@article{yang2026sceneweaver,
  title={Sceneweaver: All-in-one 3d scene synthesis with an extensible and self-reflective agent},
  author={Yang, Yandan and Jia, Baoxiong and Zhang, Shujie and Huang, Siyuan},
  journal={Advances in neural information processing systems},
  volume={38},
  pages={140319--140351},
  year={2026}
}

@article{liu2025worldcraft,
  title={Worldcraft: Photo-realistic 3d world creation and customization via llm agents},
  author={Liu, Xinhang and Tang, Chi-Keung and Tai, Yu-Wing},
  journal={arXiv preprint arXiv:2502.15601},
  year={2025}
}

@inproceedings{lin2026pat3d,
  title={Pat3d: Physics-augmented text-to-3d scene generation},
  author={Lin, Guying and Huang, Kemeng and Liu, Michael and Gao, Ruihan and Chen, Hanke and Chen, Lyuhao and Lu, Beijia and Komura, Taku and Liu, Yuan and Zhu, Jun-Yan and others},
  booktitle={International Conference on Learning Representations},
  volume={2026},
  pages={150281--150301},
  year={2026}
}

@article{huang2024story3d,
  title={Story3d-agent: Exploring 3d storytelling visualization with large language models},
  author={Huang, Yuzhou and Qin, Yiran and Lu, Shunlin and Wang, Xintao and Huang, Rui and Shan, Ying and Zhang, Ruimao},
  journal={arXiv preprint arXiv:2408.11801},
  year={2024}
}

@article{dehghanian2026camera,
  title={Camera trajectory generation: A comprehensive survey of methods, metrics, and future directions},
  author={Dehghanian, Zahra and Ardehkhani, Pouya and Vahedi, Amir and Beigy, Hamid and Rabiee, Hamid R},
  journal={Computer Vision and Image Understanding},
  pages={104872},
  year={2026},
  publisher={Elsevier}
}

@inproceedings{bahmani2025vd3d,
  title={Vd3d: Taming large video diffusion transformers for 3d camera control},
  author={Bahmani, Sherwin and Skorokhodov, Ivan and Siarohin, Aliaksandr and Menapace, Willi and Qian, Guocheng and Vasilkovsky, Michael and Lee, Hsin-Ying and Wang, Chaoyang and Zou, Jiaxu and Tagliasacchi, Andrea and others},
  booktitle={International Conference on Learning Representations},
  volume={2025},
  pages={66712--66737},
  year={2025}
}

@article{lino2015intuitive,
  title={Intuitive and efficient camera control with the toric space},
  author={Lino, Christophe and Christie, Marc},
  journal={ACM Transactions on Graphics (TOG)},
  volume={34},
  number={4},
  pages={1--12},
  year={2015},
  publisher={ACM New York, NY, USA}
}

@inproceedings{wang2024motionctrl,
  title={Motionctrl: A unified and flexible motion controller for video generation},
  author={Wang, Zhouxia and Yuan, Ziyang and Wang, Xintao and Li, Yaowei and Chen, Tianshui and Xia, Menghan and Luo, Ping and Shan, Ying},
  booktitle={ACM SIGGRAPH 2024 Conference Papers},
  pages={1--11},
  year={2024}
}

@article{he2024cameractrl,
  title={Cameractrl: Enabling camera control for text-to-video generation},
  author={He, Hao and Xu, Yinghao and Guo, Yuwei and Wetzstein, Gordon and Dai, Bo and Li, Hongsheng and Yang, Ceyuan},
  journal={arXiv preprint arXiv:2404.02101},
  year={2024}
}

@inproceedings{yang2024direct,
  title={Direct-a-video: Customized video generation with user-directed camera movement and object motion},
  author={Yang, Shiyuan and Hou, Liang and Huang, Haibin and Ma, Chongyang and Wan, Pengfei and Zhang, Di and Chen, Xiaodong and Liao, Jing},
  booktitle={ACM SIGGRAPH 2024 Conference Papers},
  pages={1--12},
  year={2024}
}

@article{xu2024camco,
  title={Camco: Camera-controllable 3d-consistent image-to-video generation},
  author={Xu, Dejia and Nie, Weili and Liu, Chao and Liu, Sifei and Kautz, Jan and Wang, Zhangyang and Vahdat, Arash},
  journal={arXiv preprint arXiv:2406.02509},
  year={2024}
}

@article{xu2024cavia,
  title={Cavia: Camera-controllable multi-view video diffusion with view-integrated attention},
  author={Xu, Dejia and Jiang, Yifan and Huang, Chen and Song, Liangchen and Gernoth, Thorsten and Cao, Liangliang and Wang, Zhangyang and Tang, Hao},
  journal={arXiv preprint arXiv:2410.10774},
  year={2024}
}

@inproceedings{bahmani2025ac3d,
  title={Ac3d: Analyzing and improving 3d camera control in video diffusion transformers},
  author={Bahmani, Sherwin and Skorokhodov, Ivan and Qian, Guocheng and Siarohin, Aliaksandr and Menapace, Willi and Tagliasacchi, Andrea and Lindell, David B and Tulyakov, Sergey},
  booktitle={2025 IEEE/CVF Conference on Computer Vision and Pattern Recognition (CVPR)},
  pages={22875--22889},
  year={2025},
  organization={IEEE}
}

@inproceedings{zhao2022particlesfm,
  title={Particlesfm: Exploiting dense point trajectories for localizing moving cameras in the wild},
  author={Zhao, Wang and Liu, Shaohui and Guo, Hengkai and Wang, Wenping and Liu, Yong-Jin},
  booktitle={European Conference on Computer Vision},
  pages={523--542},
  year={2022},
  organization={Springer}
}

@article{lin2026towards,
  title={Towards understanding camera motions in any video},
  author={Lin, Zhiqiu and Cen, Siyuan and Jiang, Daniel and Karhade, Jay and Wang, Hewei and Mitra, Chancharik and Ling, Yu Tong Tiffany and Huang, Yuhan and Zawar, Rushikesh and Bai, Xue and others},
  journal={Advances in Neural Information Processing Systems},
  volume={38},
  year={2026}
}

@inproceedings{zhang2025gendop,
  title={Gendop: Auto-regressive camera trajectory generation as a director of photography},
  author={Zhang, Mengchen and Wu, Tong and Tan, Jing and Liu, Ziwei and Wetzstein, Gordon and Lin, Dahua},
  booktitle={2025 IEEE/CVF International Conference on Computer Vision (ICCV)},
  pages={18229--18239},
  year={2025},
  organization={IEEE}
}

@article{li2024director3d,
  title={Director3d: Real-world camera trajectory and 3d scene generation from text},
  author={Li, Xinyang and Lai, Zhangyu and Xu, Linning and Qu, Yansong and Cao, Liujuan and Zhang, Shengchuan and Dai, Bo and Ji, Rongrong},
  journal={Advances in neural information processing systems},
  volume={37},
  pages={75125--75151},
  year={2024}
}

@article{xu2025filmagent,
  title={Filmagent: A multi-agent framework for end-to-end film automation in virtual 3d spaces},
  author={Xu, Zhenran and Wang, Longyue and Wang, Jifang and Li, Zhouyi and Shi, Senbao and Yang, Xue and Wang, Yiyu and Hu, Baotian and Yu, Jun and Zhang, Min},
  journal={arXiv preprint arXiv:2501.12909},
  year={2025}
}

@inproceedings{jiang2024cinematographic,
  title={Cinematographic camera diffusion model},
  author={Jiang, Hongda and Wang, Xi and Christie, Marc and Liu, Libin and Chen, Baoquan},
  booktitle={Computer Graphics Forum},
  volume={43},
  number={2},
  pages={e15055},
  year={2024},
  organization={Wiley Online Library}
}

@inproceedings{courant2024exceptional,
  title={ET the exceptional trajectories: Text-to-camera-trajectory generation with character awareness},
  author={Courant, Robin and Dufour, Nicolas and Wang, Xi and Christie, Marc and Kalogeiton, Vicky},
  booktitle={European conference on computer vision},
  pages={464--480},
  year={2024},
  organization={Springer}
}

@inproceedings{rublee2011orb,
  title={ORB: An efficient alternative to SIFT or SURF},
  author={Rublee, Ethan and Rabaud, Vincent and Konolige, Kurt and Bradski, Gary},
  booktitle={2011 International conference on computer vision},
  pages={2564--2571},
  year={2011},
  organization={Ieee}
}

@article{lin2025navcot,
  title={Navcot: Boosting llm-based vision-and-language navigation via learning disentangled reasoning},
  author={Lin, Bingqian and Nie, Yunshuang and Wei, Ziming and Chen, Jiaqi and Ma, Shikui and Han, Jianhua and Xu, Hang and Chang, Xiaojun and Liang, Xiaodan},
  journal={IEEE Transactions on Pattern Analysis and Machine Intelligence},
  volume={47},
  number={7},
  pages={5945--5957},
  year={2025},
  publisher={IEEE}
}

@misc{yolo11_ultralytics,
  author = {Jocher, Glenn and Qiu, Jing},
  title = {Ultralytics YOLO11},
  year = {2024},
  howpublished = {\url{https://github.com/ultralytics/ultralytics}},
  note = {Version 11.0.0}
}

@article{Qwen2.5-VL,
  title={Qwen2.5-VL Technical Report},
  author={Bai, Shuai and Chen, Keqin and Liu, Xuejing and Wang, Jialin and Ge, Wenbin and Song, Sibo and Dang, Kai and Wang, Peng and Wang, Shijie and Tang, Jun and Zhong, Humen and Zhu, Yuanzhi and Yang, Mingkun and Li, Zhaohai and Wan, Jianqiang and Wang, Pengfei and Ding, Wei and Fu, Zheren and Xu, Yiheng and Ye, Jiabo and Zhang, Xi and Xie, Tianbao and Cheng, Zesen and Zhang, Hang and Yang, Zhibo and Xu, Haiyang and Lin, Junyang},
  journal={arXiv preprint arXiv:2502.13923},
  year={2025}
}

@misc{google2025gemini3flash,
  author = {{Google DeepMind}},
  title = {{Gemini 3 Flash} -- Model Card},
  year = {2025},
  month = dec,
  howpublished = {Model card. \url{https://storage.googleapis.com/deepmind-media/Model-Cards/Gemini-3-Flash-Model-Card.pdf}},
  url = {https://storage.googleapis.com/deepmind-media/Model-Cards/Gemini-3-Flash-Model-Card.pdf},
  note = {API model identifier: \texttt{gemini-3-flash-preview}. Accessed 2026-09-30}
}

@article{xu2024sketch2scene,
  title={Sketch2Scene: Automatic Generation of Interactive 3D Game Scenes from User's Casual Sketches},
  author={Xu, Yongzhi and Ng, Yonhon and Wang, Yifu and Sa, Inkyu and Duan, Yunfei and Sun, Zhenhong and Li, Yang and Ji, Pan and Li, Hongdong},
  journal={arXiv preprint arXiv:2408.04567},
  year={2024}
}

@inproceedings{sun2024eggen,
  title={Eggen: Image generation with multi-entity prior learning through entity guidance},
  author={Sun, Zhenhong and Wang, Junyan and Tan, Zhiyu and Dong, Daoyi and Ma, Hailan and Li, Hao and Gong, Dong},
  booktitle={Proceedings of the 32nd ACM International Conference on Multimedia},
  pages={6637--6645},
  year={2024}
}

@article{liu2026beyond,
  title={Beyond the dirac delta: Mitigating diversity collapse in reinforcement fine-tuning for versatile image generation},
  author={Liu, Jinmei and Li, Haoru and Sun, Zhenhong and Chen, Chaofeng and Bian, Yatao and Wang, Bo and Dong, Daoyi and Chen, Chunlin and Wang, Zhi},
  journal={arXiv preprint arXiv:2601.12401},
  year={2026}
}

@article{wang2024lift,
  title={Lift: Leveraging human feedback for text-to-video model alignment},
  author={Wang, Yibin and Tan, Zhiyu and Wang, Junyan and Yang, Xiaomeng and Jin, Cheng and Li, Hao},
  journal={arXiv preprint arXiv:2412.04814},
  year={2024}
}

@article{wang2026harnesspai,
  title={HarnessPAI: An Evolving Harness for Physical AI},
  author={Wang, Xin and Wu, Wenhao and Zhang, Menghao and Wang, Zhi and Shao, Kun and Luan, Jian and Li, Yang and Li, Qing and Gu, Shangding and Zhou, Huichi and others},
  journal={arXiv preprint arXiv:2609.29166},
  year={2026}
}

@inproceedings{wu2026injection,
  title={Injection Without Distortion: Geometrically Constrained Knowledge Enhancement for Vision-Language Models},
  author={Wu, Zhongze and Su, Xiu and Yang, Feng and You, Shan and Long, Jun and Luo, Yueyi},
  booktitle={Proceedings of the AAAI Conference on Artificial Intelligence},
  volume={40},
  number={13},
  pages={10897--10905},
  year={2026}
}

@inproceedings{wu2024detecting,
  title={Detecting any instruction-to-answer interaction relationship: Universal instruction-to-answer navigator for med-vqa},
  author={Wu, Zhongze and Xu, Hongyan and Long, Yitian and You, Shan and Su, Xiu and Long, Jun and Luo, Yueyi and Xu, Chang},
  booktitle={Forty-first International Conference on Machine Learning},
  year={2024}
}

@inproceedings{hu2025divide,
  title={Divide and Conquer: Grounding {LLM}s as Efficient Decision-Making Agents via Offline Hierarchical Reinforcement Learning},
  author={Hu, Zican and Liu, Wei and Qu, Xiaoye and Yue, Xiangyu and Chen, Chunlin and Wang, Zhi and Cheng, Yu},
  booktitle={Proceedings of the 42nd International Conference on Machine Learning},
  pages={24570--24590},
  year={2025},
  volume={267},
  series={Proceedings of Machine Learning Research},
  publisher={PMLR},
  url={https://proceedings.mlr.press/v267/hu25q.html}
}

@inproceedings{wu2026marag,
    title={From Conflict to Consensus: Boosting Medical Reasoning via Multi-Round Agentic {RAG}},
    author={Wenhao Wu and Zhentao Tang and Yafu Li and Shixiong Kai and Mingxuan Yuan and Zhenhong Sun and Chunlin Chen and Zhi Wang},
    booktitle={Forty-third International Conference on Machine Learning},
    year={2026},
    url={https://openreview.net/forum?id=S7lpdz7NAW}
}

%%%%%%%%%%%%%%%%%%%%%%%%%%%%%%%%%%%%%%%%%%%%%%%%%%%%%%%%%%%%

\clearpage
\appendix
\renewcommand\mypara[1]{\vspace{1pt}\noindent\textbf{#1.}}

\newcommand{\appendixboardpair}[3]{%
\begin{figure}[H]
\centering
#1
\par\smallskip
#2
\caption{Monte Carlo viewpoint-search candidate boards.
Each board reports retained candidate views after sampling, geometric and semantic validation, deduplication, and reranking.}
\label{#3}
\end{figure}
\clearpage
}

\newcommand{\appendixboardtriple}[4]{%
\begin{figure}[H]
\centering
#1
\par\smallskip
#2
\par\smallskip
#3
\caption{Monte Carlo viewpoint-search candidate boards, pages #1--#3.
These compact boards show the final retained candidate sets before viewpoint selection.}
\label{#4}
\end{figure}
\clearpage
}

\section{Implementation Details}
\label{Appendix Implementation Detail}
\mypara{Observation contract}
For each shot, Look-Before-Move first parses the directorial intent into a semantic observation contract.
The contract specifies the primary acting subject, the expected semantic target region, visibility and occlusion requirements, composition preferences, and temporal action cue.
This contract is used to constrain the subsequent candidate search, so the system does not optimize camera motion before deciding what evidence the scene should provide.

\begin{figure}[t]
\centering
\includegraphics[width=\textwidth]{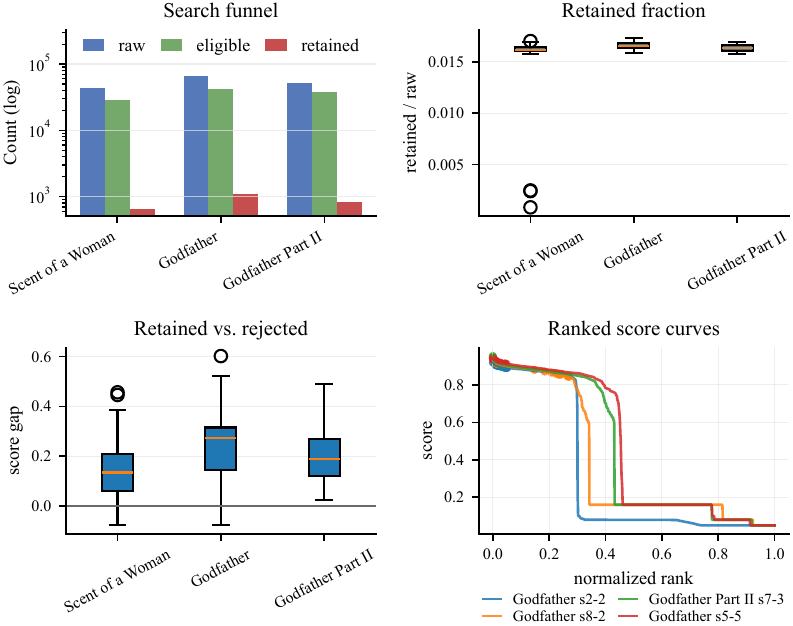}
\caption{Monte Carlo viewpoint-search summary. The search funnel, retained-fraction distribution, retained-versus-rejected score gap, and candidate ranking curve show how large raw candidate pools are filtered into compact high-quality viewpoint sets.}
\label{fig:appendix_mc_summary}
\end{figure}

\mypara{Monte Carlo viewpoint search}
The Look stage samples a large raw camera-pose pool around the current action context, rejects geometrically invalid or semantically weak views, and retains a compact candidate set for reranking.
Figure~\ref{fig:appendix_mc_summary} reports the search statistics over the available candidate-board records.
Across 132 shots with visualized candidate boards, the search produces 160,693 raw candidates, keeps 107,478 eligible candidates after geometric and semantic filtering, and retains 2,587 high-quality candidates for final reranking.
The small retained fraction indicates that the method is not selecting a viewpoint from a single heuristic anchor; it is searching a large feasible space and compressing it into a small set of narrative-compliant alternatives.

\mypara{Trajectory grounding and editing}
After viewpoint selection, the Move stage grounds discrete viewpoints into continuous camera trajectories in the executable 3D world.
Trajectory grounding follows the selected subject, avoids infeasible camera placement, and smooths motion across keyframes.
The temporal editing step then schedules observation clips by balancing local information utility against transition inconsistency, which is especially important when consecutive shots focus on different subjects or semantic targets.

\clearpage
\section{Motivation and Problem Setup}

\begin{figure}[H]
\centering
\includegraphics[width=0.96\textwidth]{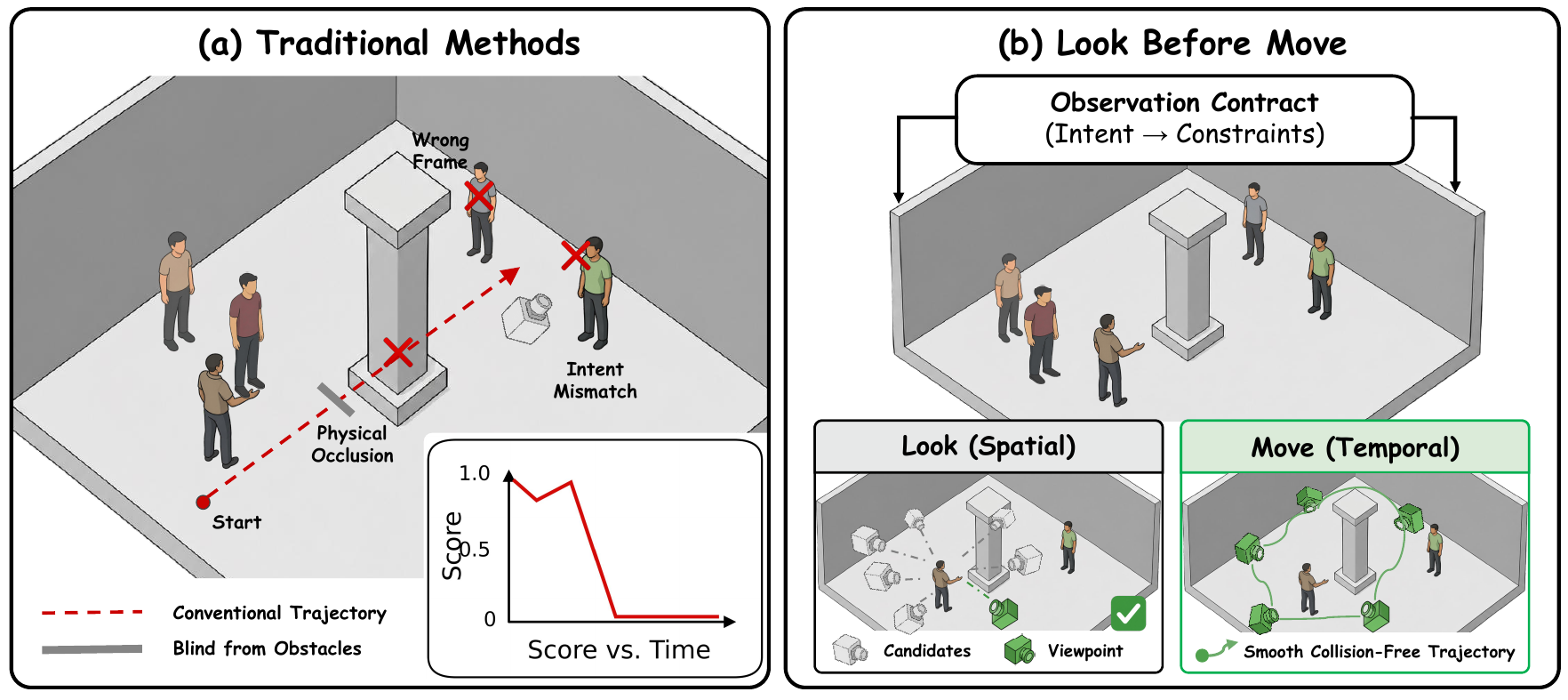}
\caption{Motivation and task contrast for narrative-grounded world visual attention. The task requires the agent to convert high-level directorial intent into observable subjects, semantic regions, feasible viewpoints, and executable camera motion in a dynamic 3D story world.}
\label{fig:appendix_motivation}
\end{figure}

Figure~\ref{fig:appendix_motivation} summarizes the motivation behind narrative-grounded world visual attention.
Traditional camera control often starts from a specified target, a coarse cinematic rule, or a desired trajectory pattern.
In contrast, camera planning inside an executable story world must first determine which visual evidence is narratively relevant: the acting subject, the semantic region, the surrounding context, and the temporal relation to adjacent events.
This difference makes the problem more than trajectory synthesis.
The camera must first look for an observation that satisfies the story intent and the physical constraints of the 3D world, and only then move through a trajectory that preserves that observation over time.

\section{Benchmark Details}
\label{Appendix Benchmark Details}
\begin{figure}[H]
\centering
\includegraphics[width=\linewidth]{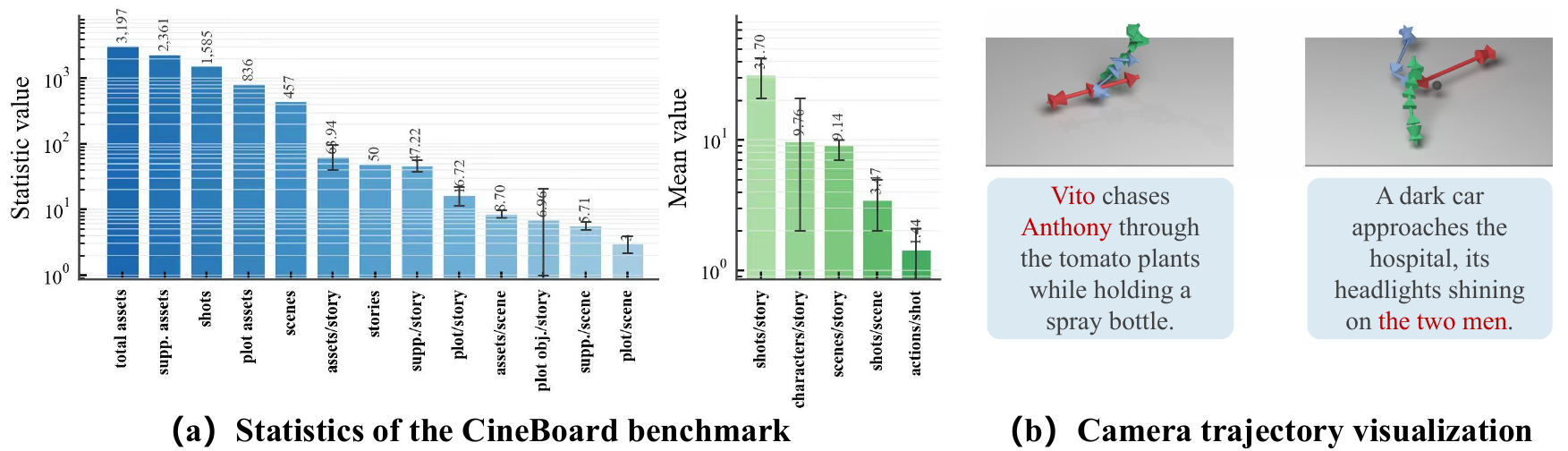}
\caption{Statistics of the 3D Story World benchmark and representative trajectory visualizations. The left panels summarize benchmark scale, asset composition, character density, and shot-level structure across executable story worlds. The right panels show how natural-language filming instructions are converted into concrete 3D camera paths for dynamic narrative events.}
\label{fig:benchmark_statistics_trajectory}
\end{figure}
\begin{table}[htbp]
\centering
\footnotesize
\renewcommand{\arraystretch}{1.10}
\caption{Metric groups used in the benchmark. The same definitions are used throughout the main and appendix results.}
\label{tab:appendix_metrics}
\setlength{\tabcolsep}{4pt}
\begin{tabular}{p{0.19\textwidth}p{0.12\textwidth}p{0.61\textwidth}}
\toprule
Group & Sub-metric & Evaluation focus \\
\midrule
Subject Perception & SP1 & Target capture and sufficient on-screen visibility of the intended subject. \\
Subject Perception & SP2 & Subject identity consistency in multi-character or visually ambiguous scenes. \\
Subject Perception & SP3 & Occlusion robustness when the subject is blocked by geometry, props, or other actors. \\
Intent Consistency & IC1 & Agreement with requested shot scale, camera angle, and composition preference. \\
Intent Consistency & IC2 & Accuracy of semantic target framing, such as face, hands, feet, or full body. \\
Intent Consistency & IC3 & Consistency with the local event description and narrative action cue. \\
Trajectory Quality & TQ1 & Motion smoothness and avoidance of abrupt camera displacement. \\
Trajectory Quality & TQ2 & Executability, stability, and collision-aware spatial behavior in the 3D scene. \\
Trajectory Quality & TQ3 & Temporal continuity between adjacent shot tails and heads. \\
\bottomrule
\end{tabular}
\end{table}

\mypara{Data hierarchy}
\label{Appendix Data hierarchy}
Our benchmark is organized as a hierarchy of stories, scenes, and shots.
Each story contains temporally ordered narrative events; each scene provides an executable 3D environment with persistent characters, objects, layout, and animations; each shot specifies the local acting subject, semantic target, action cue, and directorial requirement that the camera must satisfy.
The benchmark contains 50 stories, 457 scenes, and 1585 shots, with 3197 explicit 3D assets including 836 plot-related assets and 2361 supplementary scene assets.
This structure allows us to evaluate camera plans with closed-loop rendering feedback rather than treating viewpoint prediction as a static pose-regression task.

\mypara{Evaluation dimensions}
\label{Appendix Evaluation dimensions}
We evaluate all methods with the same rendered evidence, evaluator models, and segment-weighted aggregation.
The metrics are grouped into Subject Perception (SP), Intent Consistency (IC), and Trajectory Quality (TQ), all normalized to a 0--100 scale.
SP measures whether the intended subject remains visible, correctly identified, and robust to occlusion.
IC measures whether the rendered view follows the requested shot scale, semantic target, and event description.
TQ measures whether the final camera motion is smooth, stable, and temporally coherent across adjacent shots.

\mypara{Metric computation}
\label{Appendix Metric computation}
For each method, we render the planned trajectory into a sequence of frames.
Let shot $i$ contain $T_i$ frames $\{F_{i,t}\}_{t=1}^{T_i}$ and let the observation contract specify the target subject $a_i^*$, semantic target $z_i$, shot-size request $q_i$, and event description $e_i$.
The evaluator extracts a target box $b_{i,t}$ with confidence $c_{i,t}$, area ratio $r_{i,t}=|b_{i,t}|/|F_{i,t}|$, normalized center distance $d_{i,t}$, and a validity flag $m_{i,t}$ indicating whether the intended target is detected.
We use $\operatorname{clip}(x)=\min(1,\max(0,x))$ and map every sub-score to 0--100.
Invalid or inapplicable terms are removed from the denominator; in particular, TQ3 is not computed for the first shot of a story segment.

\begin{align}
\mathrm{SP1}_i
&= \frac{100}{\sum_t m_{i,t}+\epsilon}
   \sum_{t=1}^{T_i} m_{i,t}\,
   c_{i,t}\,\operatorname{clip}\!\left(\frac{r_{i,t}}{\rho_{\min}}\right)
   (1-d_{i,t}), \\
\mathrm{SP2}_i
&= \frac{100}{\sum_t m_{i,t}+\epsilon}
   \sum_{t=1}^{T_i} m_{i,t}\,
   \operatorname{clip}\!\left(
   \frac{M(b_{i,t},G(a_i^*))-\max_{a\neq a_i^*}M(b_{i,t},G(a))}
        {\gamma_{\mathrm{id}}}
   \right), \\
\mathrm{SP3}_i
&=100\,\operatorname{clip}\!\left(
1-\lambda_m L_i^{\mathrm{miss}}
 -\lambda_e L_i^{\mathrm{edge}}
 -\lambda_a \frac{1}{T_i-1}\sum_{t=2}^{T_i}
     \left|\log\frac{r_{i,t}+\epsilon}{r_{i,t-1}+\epsilon}\right|
\right).
\end{align}

Here, SP1 measures effective subject coverage by combining detection confidence, visible area, and centering.
SP2 measures identity consistency using ORB feature matching $M(\cdot,\cdot)$ between the detected subject crop and the reference gallery $G(a)$ for each character; it rewards a positive margin over visually competing identities.
SP3 penalizes target disappearance $L_i^{\mathrm{miss}}$, edge truncation $L_i^{\mathrm{edge}}$, and abrupt visible-area changes, which are common symptoms of occlusion or unstable framing.

\begin{align}
\mathrm{IC1}_i
&= \frac{100}{\sum_t m_{i,t}+\epsilon}
   \sum_{t=1}^{T_i} m_{i,t}\,
   \operatorname{clip}\!\left(
   1-\frac{\left|\log\frac{r_{i,t}+\epsilon}{\rho(q_i)+\epsilon}\right|}
          {\tau_{\mathrm{size}}}
   \right), \\
\mathrm{IC2}_i
&= \frac{100}{\sum_t m_{i,t}+\epsilon}
   \sum_{t=1}^{T_i} m_{i,t}\, S_{\mathrm{region}}(F_{i,t}, b_{i,t}, z_i), \\
\mathrm{IC3}_i
&=100\, S_{\mathrm{vlm}}\!\left(e_i,\{F_{i,t}\}_{t\in \mathcal{K}_i}\right).
\end{align}

IC1 compares the observed subject area ratio with the canonical ratio $\rho(q_i)$ for the requested shot size.
IC2 evaluates whether the requested semantic region, such as face, hands, feet, or full body, is actually emphasized inside the target crop.
IC3 is a blind visual-language judgment over keyframes $\mathcal{K}_i$, where the evaluator checks whether the rendered evidence conveys the event description and directorial intent.

\begin{align}
\mathrm{TQ1}_i
&=100\,\operatorname{clip}\!\left(
1-\alpha_f \overline{\Delta \phi}_i
 -\alpha_p \overline{\mathrm{MAD}}_i
\right), \\
\mathrm{TQ2}_i
&=100\,\operatorname{clip}\!\left(
1-\beta_c \frac{1}{T_i-1}\sum_{t=2}^{T_i}\|p_{i,t}-p_{i,t-1}\|_2
 -\beta_r \frac{1}{T_i-1}\sum_{t=2}^{T_i}
     \left|\log\frac{r_{i,t}+\epsilon}{r_{i,t-1}+\epsilon}\right|
 -\beta_o C_i
\right), \\
\mathrm{TQ3}_i
&=100\,\operatorname{clip}\!\left(
\eta_s\,\mathrm{sim}(F_{i-1,T_{i-1}},F_{i,1})
\,+\,\eta_p\!\left(1-\frac{\|p_{i-1,T_{i-1}}-p_{i,1}\|_2}{\tau_{\mathrm{cut}}}\right)
\right).
\end{align}

TQ1 measures visual smoothness using Farneback optical-flow acceleration $\overline{\Delta \phi}_i$ and mean absolute pixel difference $\overline{\mathrm{MAD}}_i$.
TQ2 measures tracking stability through target-center drift $p_{i,t}$, target-area jumps, and an execution penalty $C_i$ for collisions or invalid camera placements.
TQ3 measures cut continuity between adjacent shots by combining image similarity and the jump of the target center from the previous tail frame to the current head frame.
The final group scores are simple means over valid sub-metrics,
\begin{align}
\mathrm{SP}_i
&=\frac{\mathrm{SP1}_i+\mathrm{SP2}_i+\mathrm{SP3}_i}{3}, \\
\mathrm{IC}_i
&=\frac{\mathrm{IC1}_i+\mathrm{IC2}_i+\mathrm{IC3}_i}{3}, \\
\mathrm{TQ}_i
&=\frac{\mathrm{TQ1}_i+\mathrm{TQ2}_i+\nu_i\mathrm{TQ3}_i}{2+\nu_i},
\qquad \nu_i=\mathbb{1}[i>1].
\end{align}
and the shot-level overall score is
\begin{equation}
\mathrm{Overall}_i=\frac{\mathrm{SP}_i+\mathrm{IC}_i+\mathrm{TQ}_i}{3}.
\end{equation}
For a method $m$, we report segment-count weighted results over story segments $\mathcal{S}$:
\begin{equation}
\mathrm{Score}(m)=
\frac{\sum_{s\in\mathcal{S}} n_s \left(\frac{1}{n_s}\sum_{i\in s}\mathrm{Score}_i(m)\right)}
     {\sum_{s\in\mathcal{S}} n_s},
\end{equation}
where $n_s$ is the number of evaluated shots in segment $s$.
The same aggregation is used for every sub-metric, group mean, and overall score in the main paper.

\section{Additional Ablation Analysis}

\begin{figure*}[t]
\centering
\includegraphics[width=\textwidth]{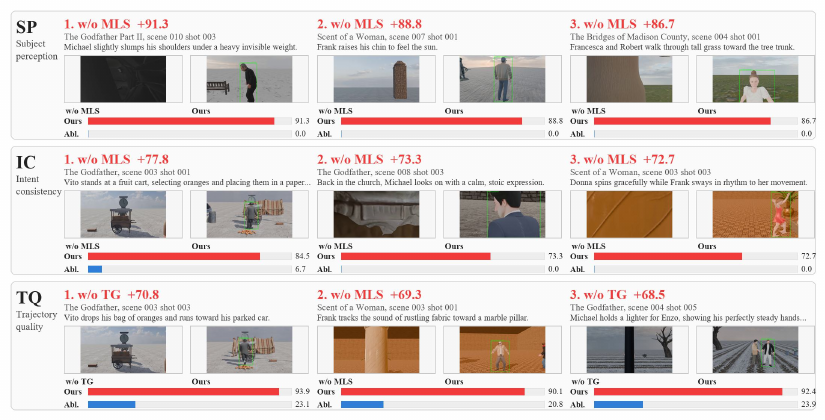}
\caption{Counterfactual ablation visualization. The metric panels show how each component affects SP, IC, and TQ sub-metrics under controlled component removals.}
\label{fig:appendix_ablation}
\end{figure*}

Figure~\ref{fig:appendix_ablation} expands the ablation table in the main paper with fine-grained sub-metric behavior and pairwise method similarity.
Multi-level Monte Carlo Search (MLS) mainly enlarges the feasible viewpoint set and improves both subject perception and intent consistency.
VLM Reflection (VLM-R) provides targeted semantic correction, especially for cases where the target region or narrative focus is visually subtle.
Trajectory Grounding (TG) turns selected viewpoints into executable motion, Semantic Height Adjust (SHA) improves the vertical alignment of semantic targets such as face or full body, and Pre-continuity Story Judge (PCSJ) reduces temporally inconsistent viewpoint switches before final editing.
The counterfactual cases further show that the components fail in different ways rather than contributing an interchangeable score boost.
When the search component is weakened, the planner often keeps a plausible subject in view but loses the best semantic region or chooses a viewpoint with fragile visibility.
When reflection is removed, errors are more likely to appear as fine-grained semantic mismatches, because the candidate view may be geometrically valid while still emphasizing the wrong body part, object, or interaction cue.
When trajectory grounding is removed, the initial still view can remain reasonable, but the rendered clip becomes unstable once the subject or camera begins to move.
This separation is useful because it indicates which part of the pipeline should be improved for each error type.

\section{Additional Qualitative Results}

The qualitative comparisons in the main paper show representative final renderings, while the candidate boards in Figures~\ref{fig:appendix_mc_boards_01_02}--\ref{fig:appendix_mc_boards_11_12} expose the intermediate viewpoint alternatives considered before final selection.
These boards are useful because many sampled views can satisfy a coarse visibility constraint, but only a smaller subset simultaneously preserves subject identity, semantic target region, action cue, occlusion constraints, and composition.
Each page corresponds to a group of shots and shows the retained candidate set after geometric validation, semantic filtering, deduplication, and reranking.

The boards also make the role of the Look stage visually explicit.
For close-up or region-specific requests, the retained views often concentrate around the requested semantic region while still keeping enough body or scene context for identity and action understanding.
For wide or motion-oriented requests, the retained views spread over a larger feasible region so that the Move stage can later choose a trajectory that is both stable and narratively informative.
This intermediate evidence helps distinguish Look-Before-Move from methods that directly commit to a single camera anchor or trajectory template.
The additional boards also clarify how the same narrative instruction can admit several visually acceptable solutions.
Some candidates prioritize a stable view of the acting subject, while others reveal more context or preserve a clearer action cue.
Keeping these alternatives until reranking gives the planner room to trade off subject scale, semantic target visibility, occlusion risk, and trajectory continuity.
This is particularly important for dynamic scenes, where a viewpoint that is optimal at one frame may become unreliable a few frames later.
The board evidence therefore complements the final renderings by showing not only what was selected, but what competing observations were rejected.

\clearpage
\noindent\begin{minipage}{\linewidth}
\section{Discussion}
\label{app:discuss}

\begin{figure}[H]
\centering
\includegraphics[width=\textwidth]{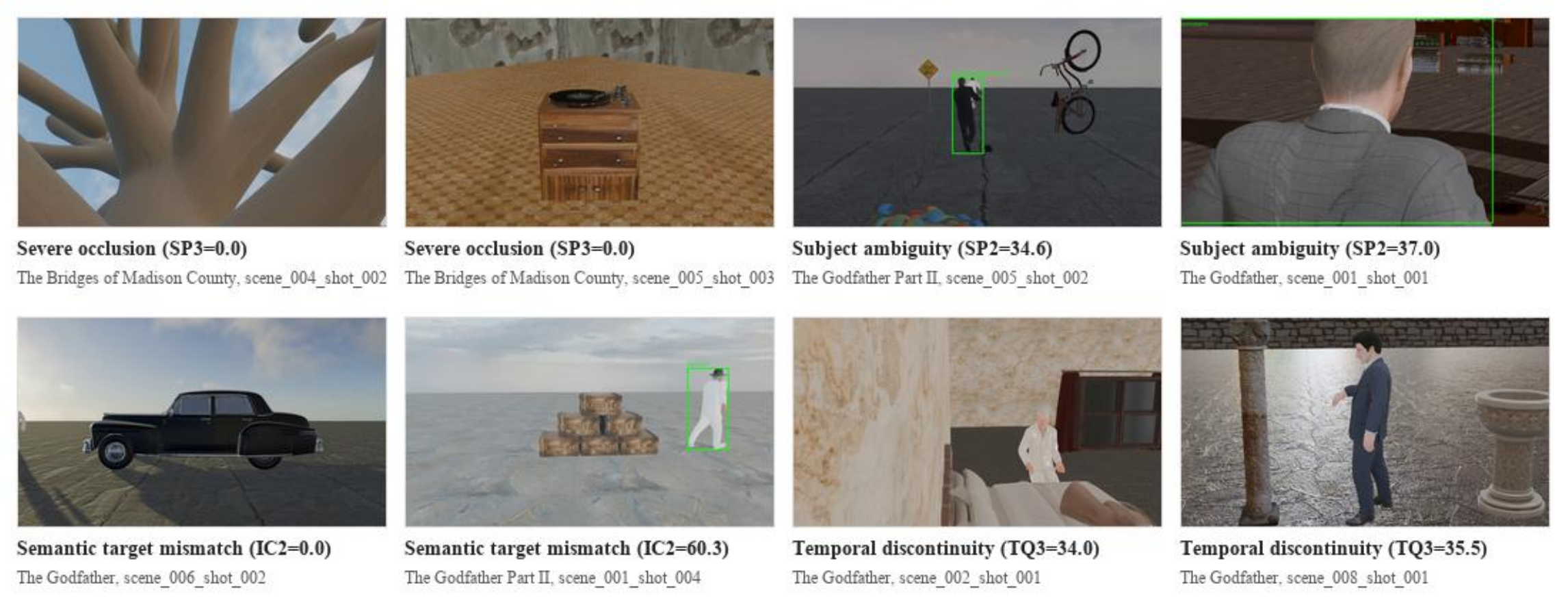}
\caption{Representative failure cases. Severe occlusion, subject ambiguity, semantic target mismatch, and temporal discontinuity correspond to drops in SP3, SP2, IC2, and TQ3, respectively.}
\label{fig:appendix_failures}
\end{figure}
\end{minipage}

\mypara{Failure Cases}
Figure~\ref{fig:appendix_failures} summarizes representative failure cases according to the most affected metric dimension. 
Severe occlusion reduces SP when the acting subject is blocked by walls, furniture, props, or other characters. 
Subject ambiguity reduces SP when the camera locks onto the wrong person or fails to maintain identity in crowded scenes. 
Semantic target mismatch reduces IC when the requested visual region, such as the face, hands, feet, or full body, is framed imprecisely. 
Temporal discontinuity reduces TQ when the visual state at the end of one shot is poorly aligned with the beginning of the next shot. 
These failures are often coupled: an early identity error can lead to later target-region mismatch, while a correct still viewpoint may become invalid once the subject moves behind geometry or another character enters the line of sight. 
This motivates reporting failures with affected sub-metrics rather than treating them as generic visual artifacts.

\mypara{Future Work}
The observed failures suggest several directions for improvement. 
First, future planners should maintain stronger scene-level reasoning and uncertainty estimates over target identity, especially in multi-character scenes. 
Second, viewpoint search should anticipate future occlusions rather than only repairing them after rendering, enabling more reliable long-horizon camera motion. 
Third, semantic targets should be treated as temporally extended visual evidence rather than local keyframe regions. 
For example, a face, hand, or full-body request may be visible at one keyframe but become less legible during camera transition. 
A stronger planner should therefore evaluate whether the requested evidence remains interpretable across the whole clip. 
In addition, future systems could return structured failure explanations together with rendered clips, allowing the camera agent to revise the specific failed constraint instead of repeating the entire planning process. 
Longer story arcs also introduce higher-level constraints such as screen direction, repeated object viewpoints, delayed reveals, and story-level visual memory, which remain important extensions of narrative-grounded world visual attention.

\mypara{Societal Impact}
This work may have positive impact by enabling more controllable and interpretable camera planning for virtual production, animation, embodied AI, simulation, and educational storytelling. 
By grounding camera behavior in explicit narrative intent and executable 3D verification, the proposed framework can reduce manual effort in scene visualization and make cinematic content creation more accessible to users without advanced technical skills. 
At the same time, more autonomous camera planning systems may be misused to generate misleading, manipulative, or low-quality synthetic visual narratives at scale. 
Therefore, future deployment should include human oversight, provenance tracking, and appropriate safeguards when the generated camera behavior is used in public-facing media or high-stakes simulation contexts.

\noindent\begin{minipage}{\linewidth}
\section{User Study}
\label{Appendix User Study}
We conducted a subjective user study to assess the perceptual quality from a human perspective. A total of 95 participants rated camera motion clips generated by Look-Before-Move and other methods (CCD, GenDoP, and Director) across four different stories, with three clips per story presented. To reduce prior bias, amera motion generation method names were hidden from participants during evaluation; within each story, clips from different methods were presented in randomized order as Clip 1--3. Participants rated each clip on three dimensions: subject perception, intent consistency, and trajectory quality, using an integer scale from 0 (extremely poor) to 5 (excellent). The final score per method was averaged across all participants and clips.

\begin{figure}[H]
\centering
\includegraphics[height=0.74\textheight]{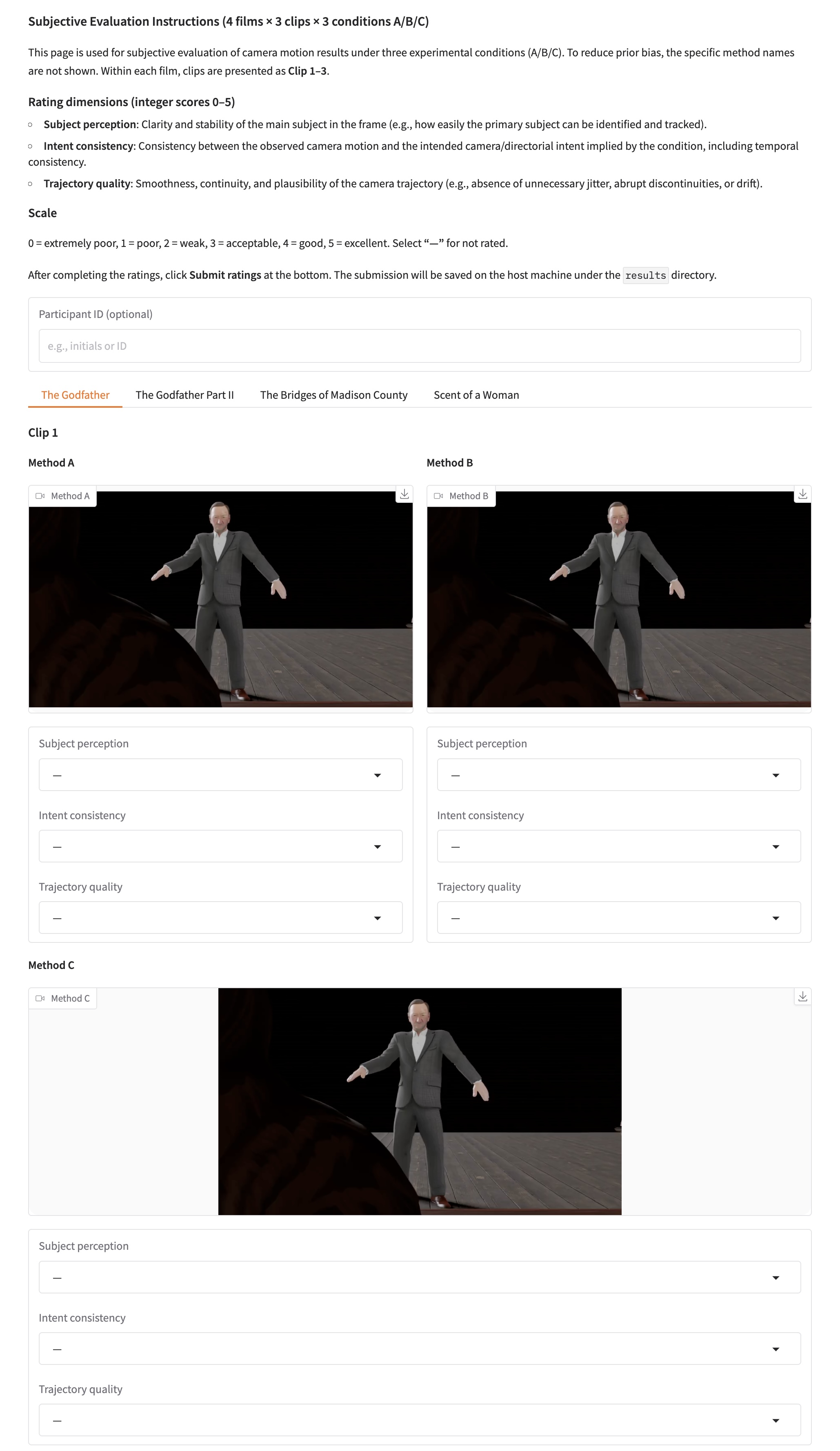}
\caption{User study interface}
\label{fig:user_study_interface}
\end{figure}
\end{minipage}

\clearpage
\appendixboardpair
{\includegraphics[width=\textwidth,height=0.47\textheight,keepaspectratio]{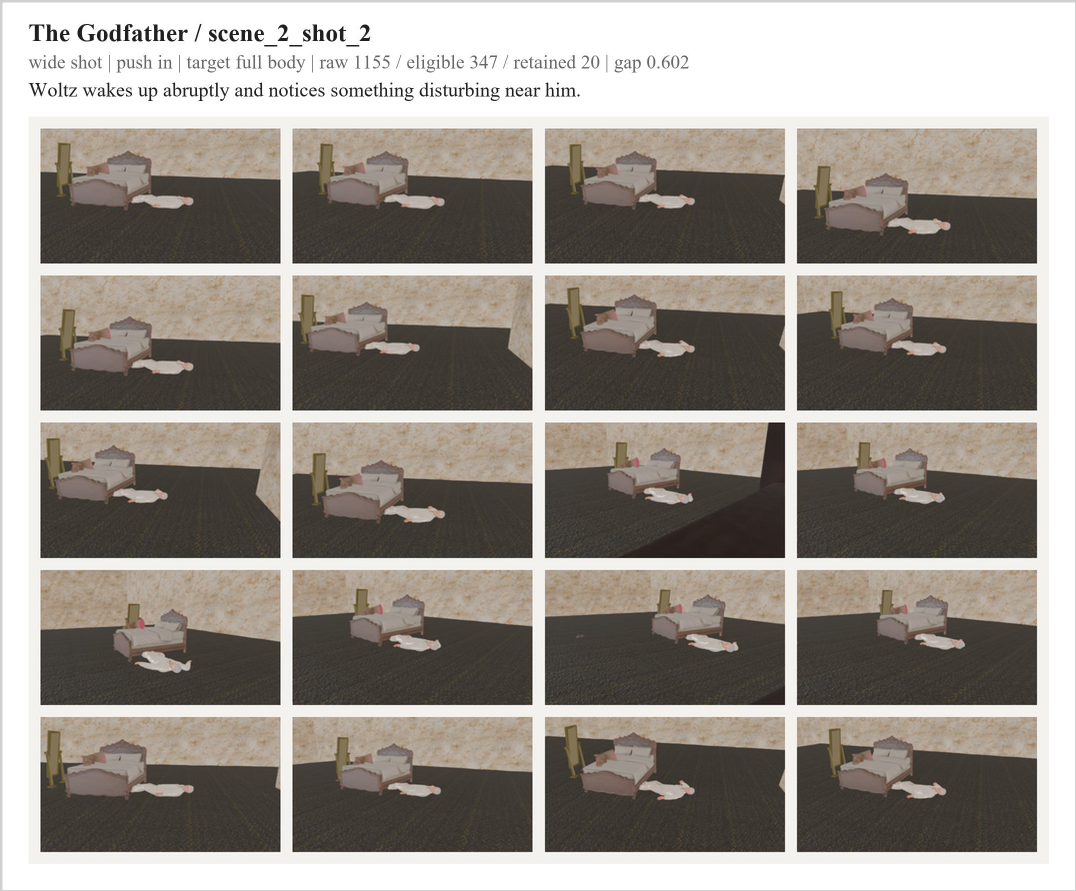}}
{\includegraphics[width=\textwidth,height=0.47\textheight,keepaspectratio]{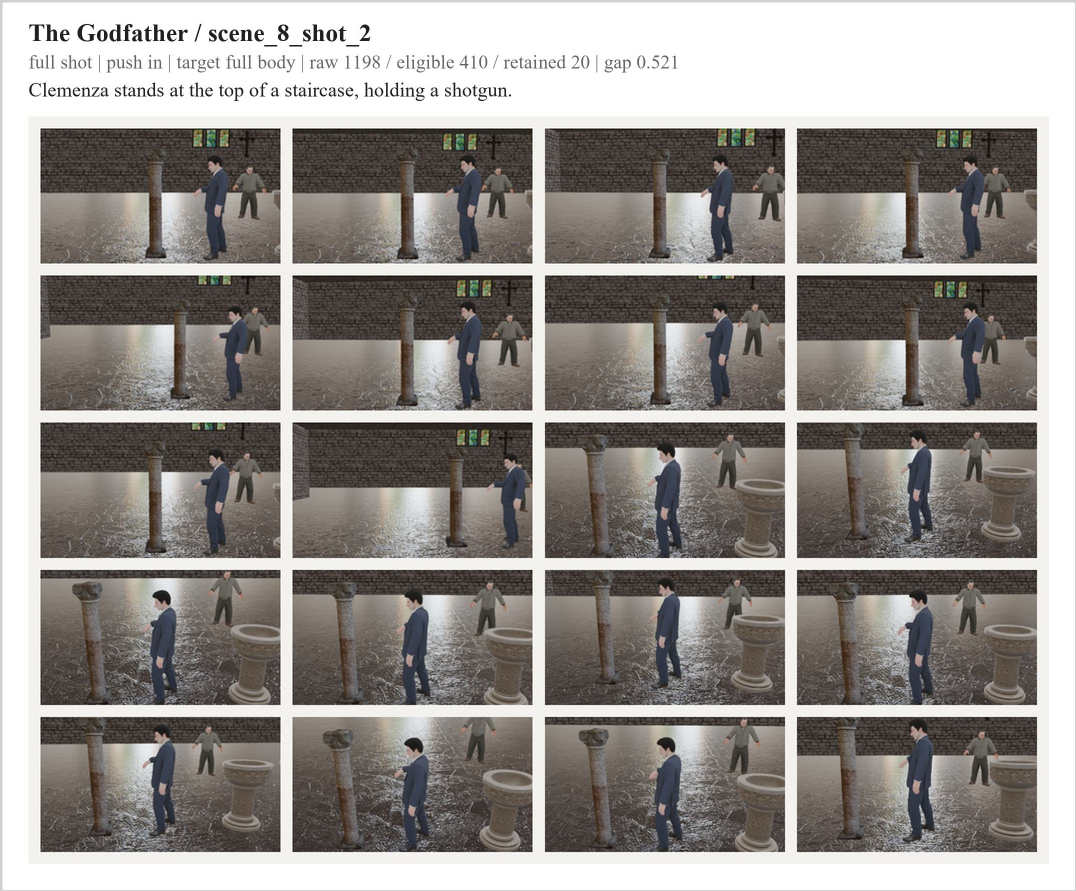}}
{fig:appendix_mc_boards_01_02}

\appendixboardpair
{\includegraphics[width=\textwidth,height=0.47\textheight,keepaspectratio]{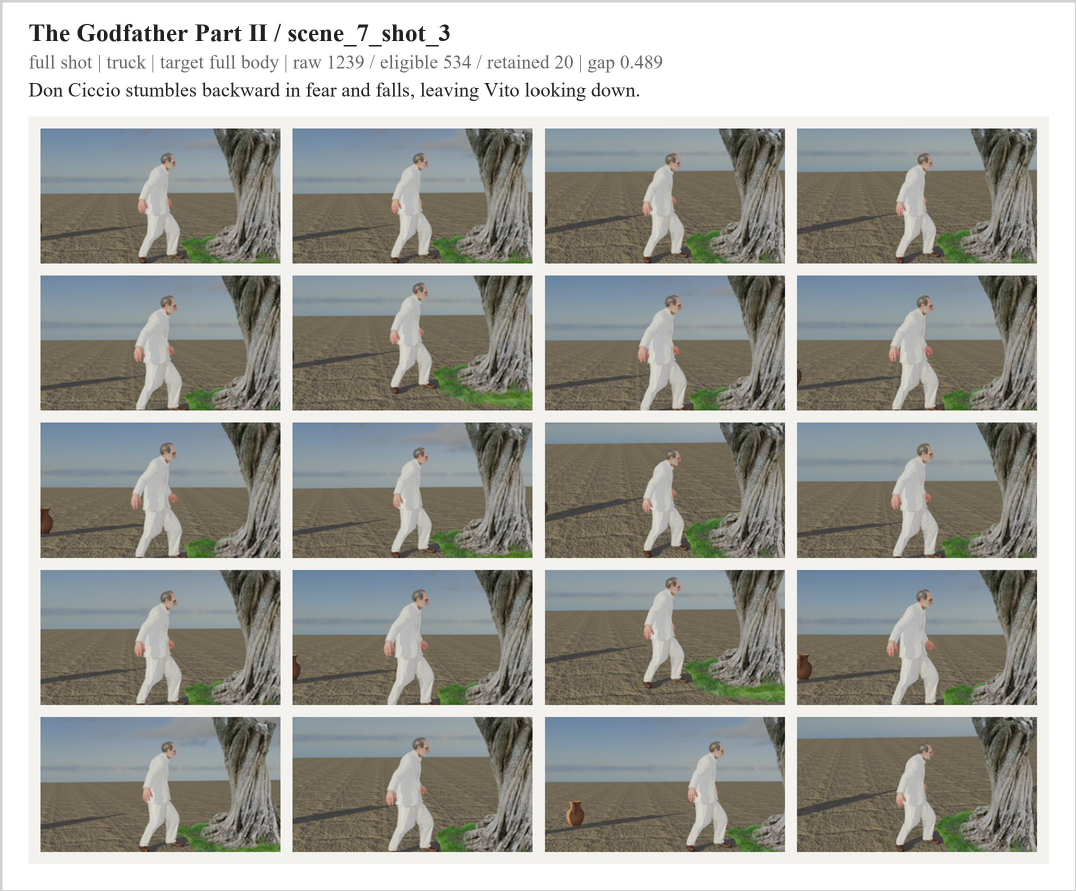}}
{\includegraphics[width=\textwidth,height=0.47\textheight,keepaspectratio]{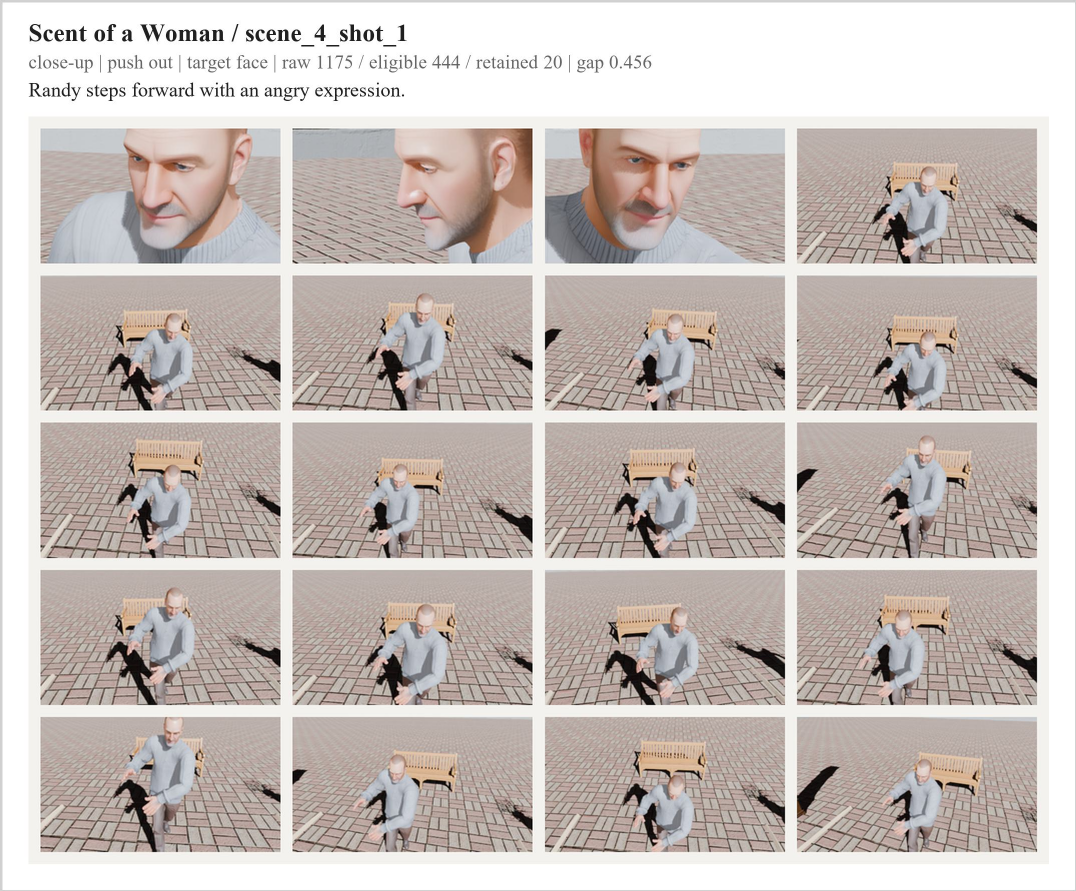}}
{fig:appendix_mc_boards_03_04}

\appendixboardpair
{\includegraphics[width=\textwidth,height=0.47\textheight,keepaspectratio]{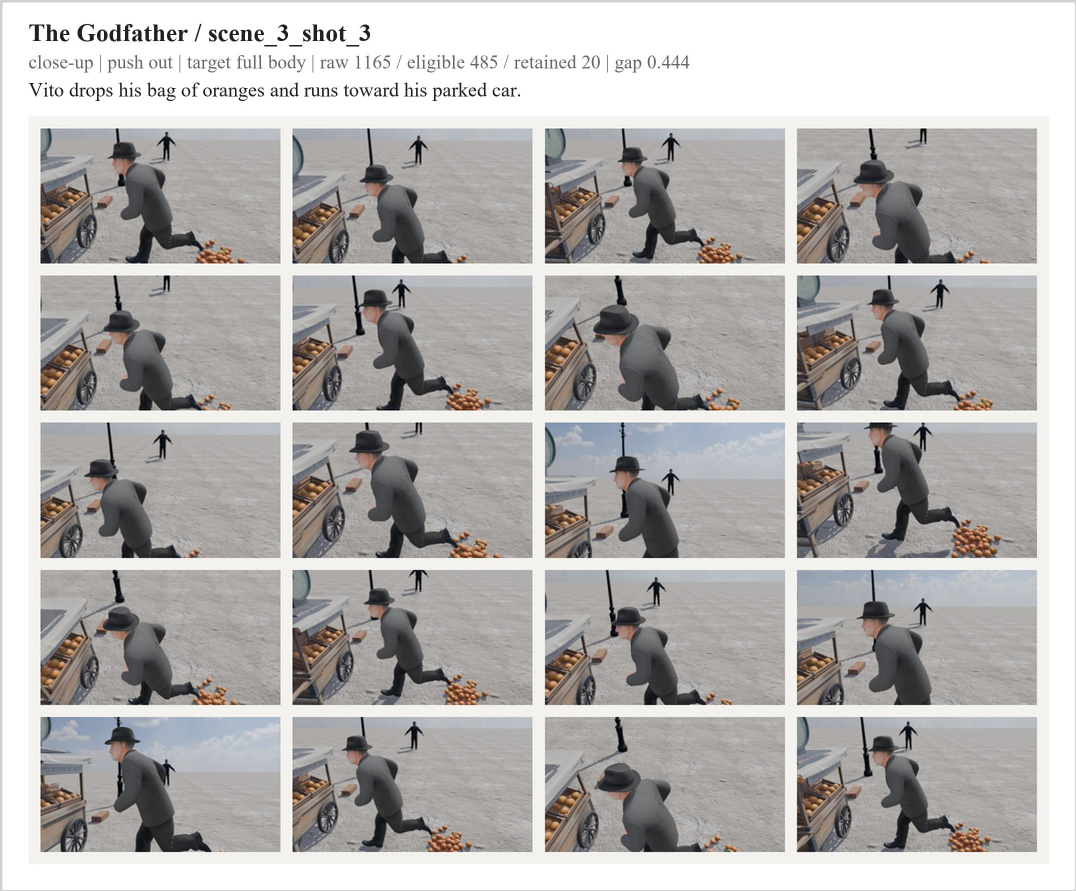}}
{\includegraphics[width=\textwidth,height=0.47\textheight,keepaspectratio]{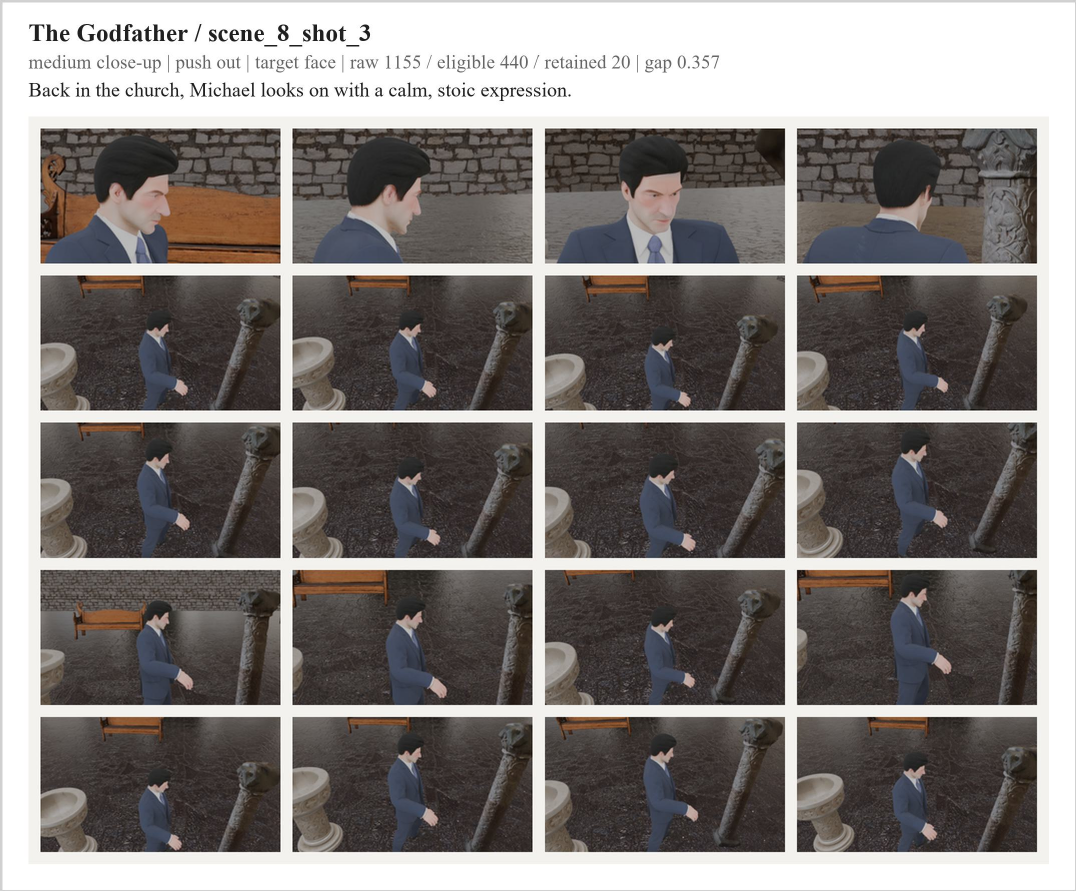}}
{fig:appendix_mc_boards_05_06}

\appendixboardpair
{\includegraphics[width=\textwidth,height=0.47\textheight,keepaspectratio]{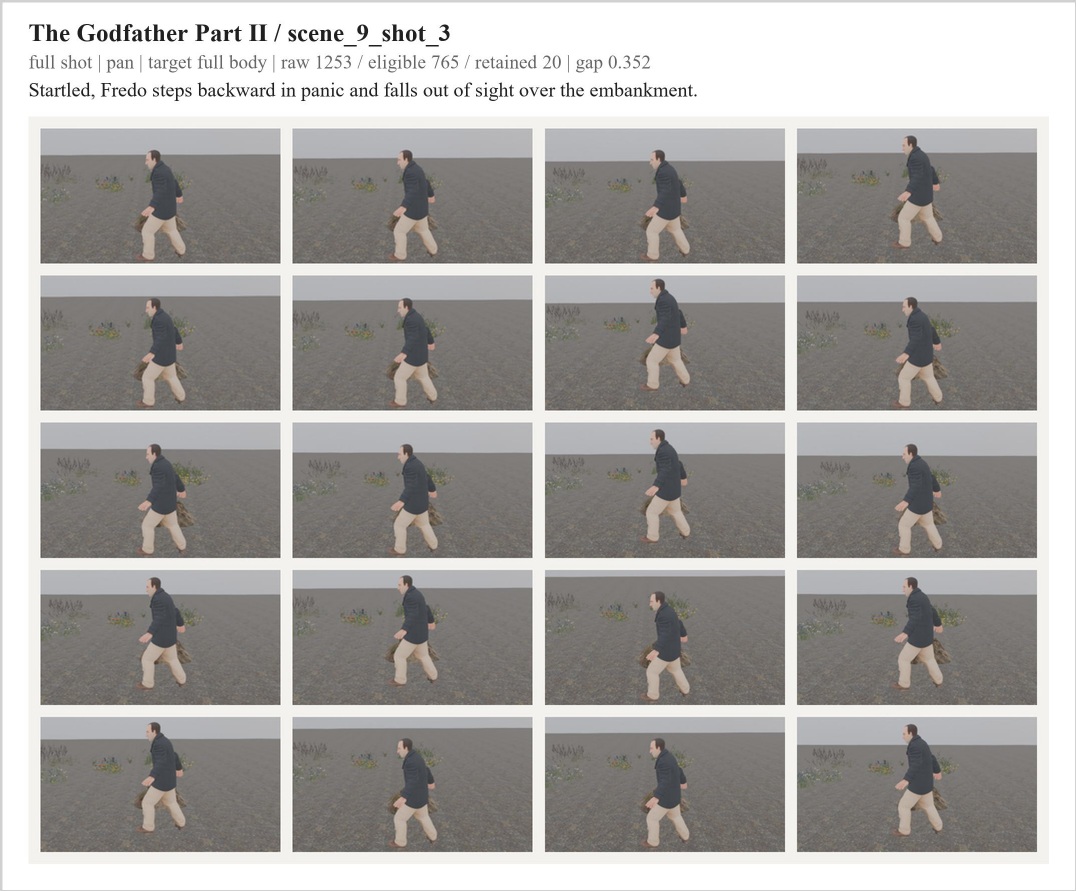}}
{\includegraphics[width=\textwidth,height=0.47\textheight,keepaspectratio]{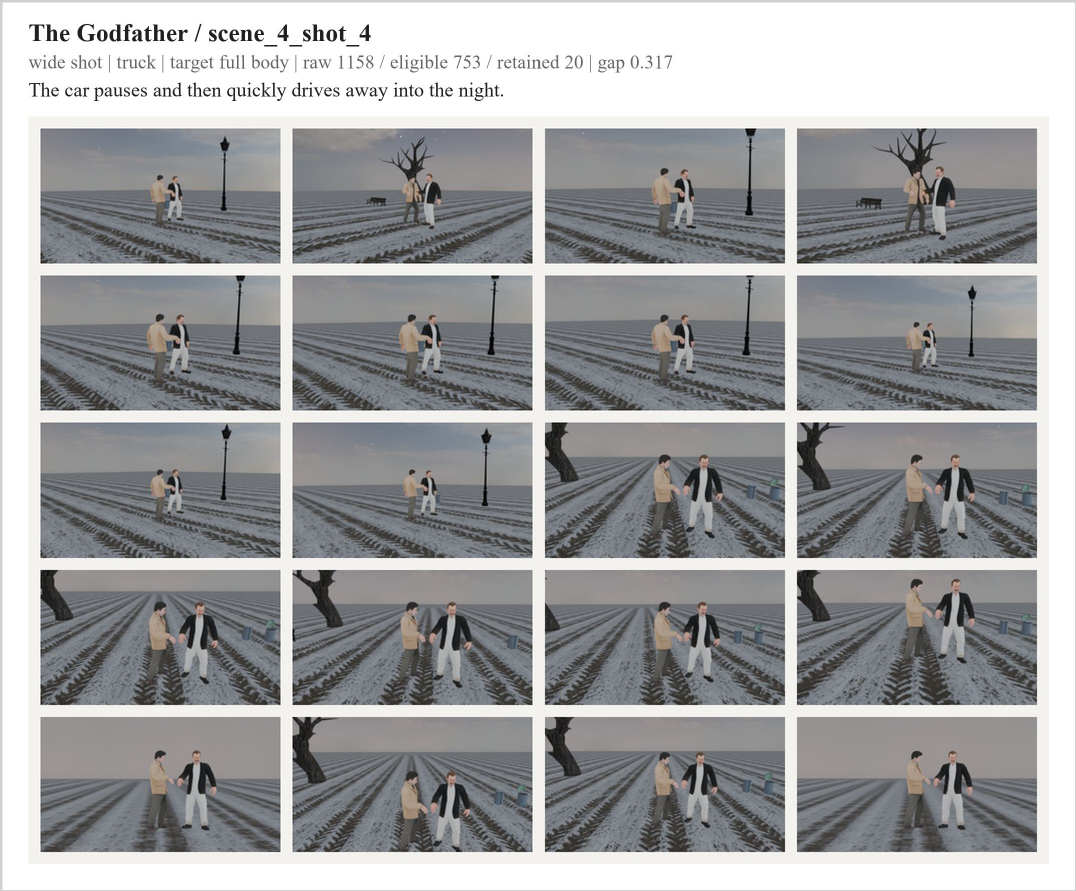}}
{fig:appendix_mc_boards_07_08}

\appendixboardpair
{\includegraphics[width=\textwidth,height=0.47\textheight,keepaspectratio]{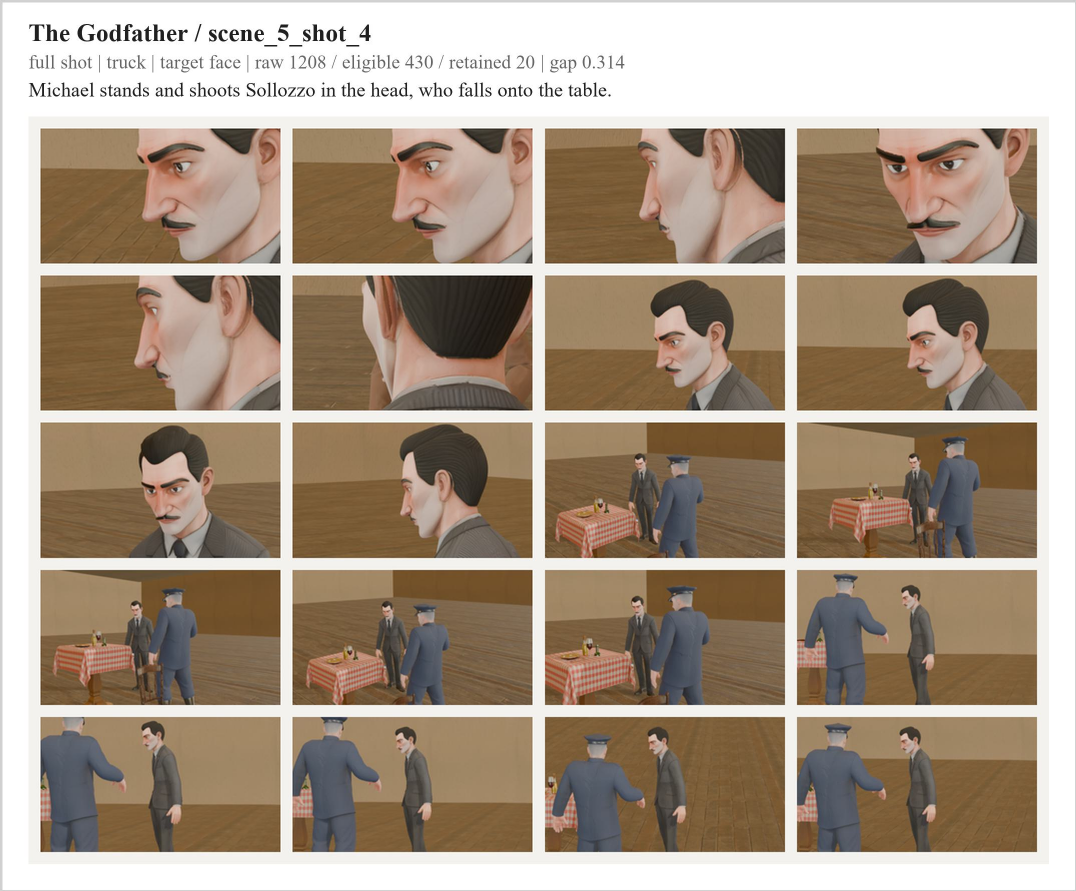}}
{\includegraphics[width=\textwidth,height=0.47\textheight,keepaspectratio]{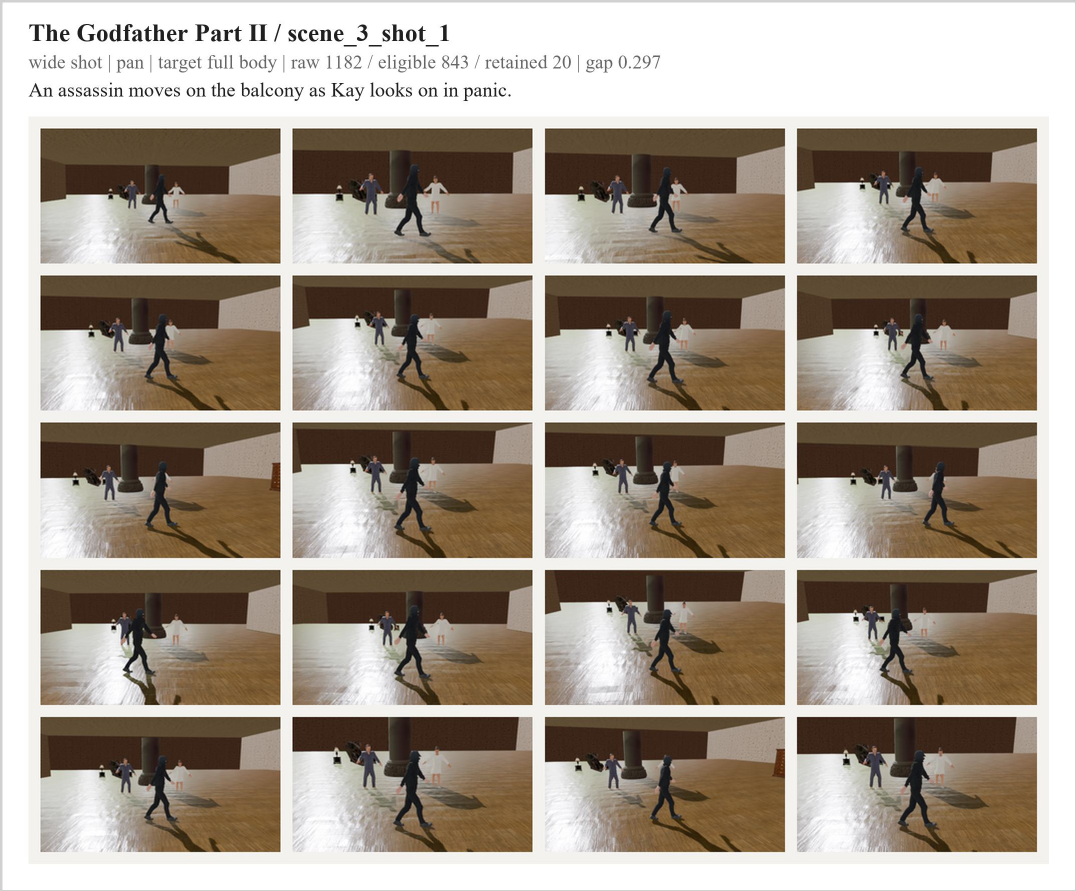}}
{fig:appendix_mc_boards_09_10}

\appendixboardpair
{\includegraphics[width=\textwidth,height=0.47\textheight,keepaspectratio]{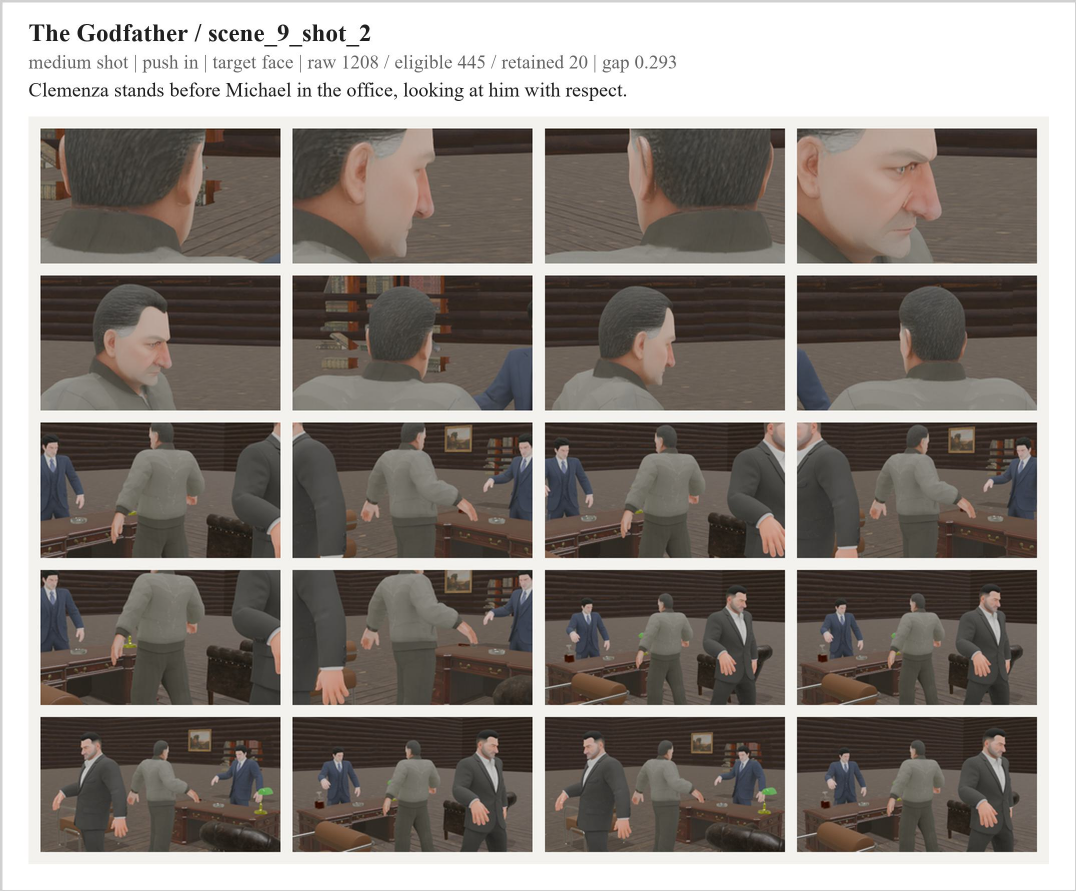}}
{\includegraphics[width=\textwidth,height=0.47\textheight,keepaspectratio]{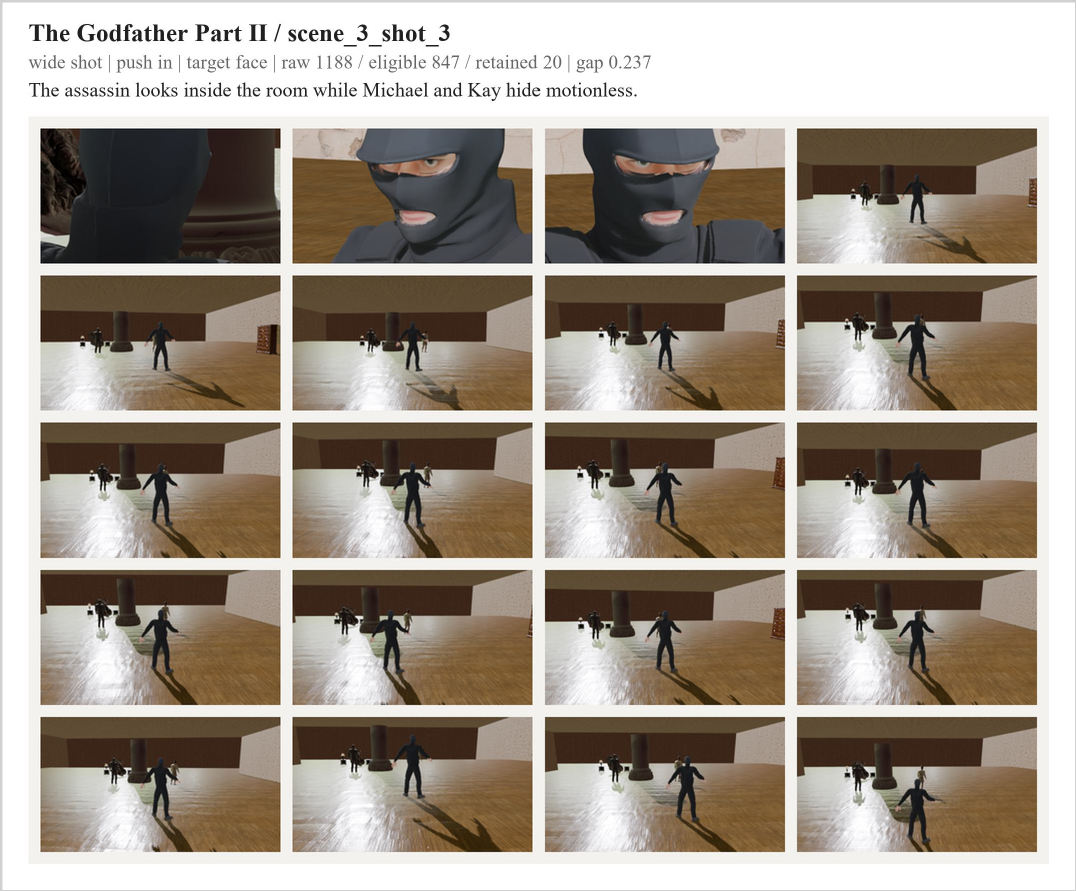}}
{fig:appendix_mc_boards_11_12}

%%%%%%%%%%%%%%%%%%%%%%%%%%%%%%%%%%%%%%%%%%%%%%%%%%%%%%%%%%%%

%%%%%%%%%%%%%%%%%%%%%%%%%%%%%%%%%%%%%%%%%%%%%%%%%%%%%%%%%%%%
\clearpage
% Appendix fragment; scientific content and values are taken from the complete
% Desktop rebuttal.tex (2026-09-30), without changing the original paper.
% Requires booktabs, already present in the original paper preamble.

\section{Observation Contract and Candidate-Budget Analyses}
\label{app:cr_contract_budget}

\subsection{Observation Contract and Direct Trajectory Generation}
\label{app:cr_contract}

The Observation Contract converts each instruction into explicit requirements for the target subject, composition, semantic region, and action cue before trajectory generation.
To isolate its contribution, we compare Direct Trajectory Generation, Observation Contract + Trajectory Generation, and LoBeM-full under identical inputs, model settings, rendering, and evaluation.
The first two use the same trajectory generator and differ only in whether the Observation Contract is used.

\begin{table}[htbp]
\centering
\small
\setlength{\tabcolsep}{3pt}
\renewcommand{\arraystretch}{1.12}
\caption{Controlled comparison of direct trajectory generation, Observation-Contract-conditioned generation, and the full framework.}
\label{app:cr_contract_table}
\resizebox{\linewidth}{!}{%
\begin{tabular}{@{}lrrrr@{}}
\toprule
Method & SP $\uparrow$ & IC $\uparrow$ & TQ $\uparrow$ & Overall $\uparrow$ \\
\midrule
LoBeM-full & $76.53\pm14.23$ & $63.90\pm17.79$ & $74.65\pm15.55$ & $\mathbf{71.70\pm11.94}$ \\
Observation Contract + Trajectory Generation & $71.32\pm23.76$ & $60.63\pm22.51$ & $63.49\pm16.68$ & $\mathbf{65.15\pm18.41}$ \\
Direct Trajectory Generation & $67.19\pm23.46$ & $53.28\pm21.12$ & $64.91\pm15.79$ & $\mathbf{61.79\pm16.40}$ \\
\bottomrule
\end{tabular}
}
\end{table}

Compared with direct generation, the Observation Contract improves Overall by 3.36 points, SP by 4.13 points, and IC by 7.35 points, isolating the benefit of structuring what should be observed before motion planning.
LoBeM-full further improves Overall by 6.55 points, demonstrating the aggregate benefits of search, geometric verification, grounding, and reflection.

\subsection{Candidate-Budget Analysis}
\label{app:cr_budget}

We disabled LLM-based reranking and reflection and varied only the number of sampled candidates.

\begin{table}[htbp]
\centering
\small
\setlength{\tabcolsep}{5pt}
\caption{Candidate-budget analysis with LLM-based reranking and reflection disabled. Candidates denotes the number of sampled candidates per camera.}
\label{app:cr_budget_table}
\begin{tabular}{@{}crrr@{}}
\toprule
Candidates & Overall $\uparrow$ & IC $\uparrow$ & Search time (s/shot) $\downarrow$ \\
\midrule
8 & 66.719 & 53.868 & 9.644 \\
16 & 69.917 & 55.354 & 10.662 \\
32 & 70.064 & 55.445 & 12.624 \\
50 & \textbf{70.319} & \textbf{56.271} & 13.710 \\
\bottomrule
\end{tabular}
\end{table}

Performance rises from 8 to 16 candidates and largely plateaus over 16--50: 16$\rightarrow$50 gains only 0.402 Overall and 0.917 IC while search time rises 28.6\%.
Thus, the full-framework gain is not explained by pose-pool size alone.

\section{Independent IC3 Evaluation}
\label{app:cr_independent_ic3}

We recomputed IC3 on the complete test set using GPT-4o, Gemini-2.5-Flash, and Claude-Sonnet-4.
All evaluators received the same keyframes, event descriptions, scoring prompt, and required output format.
Therefore, the evaluator model was the only experimental variable.

\begin{table}[htbp]
\centering
\small
\setlength{\tabcolsep}{4pt}
\caption{IC3 results under independent evaluator models and the original Qwen evaluator. Gemini, Claude, and Qwen denote Gemini-2.5-Flash, Claude-Sonnet-4, and Qwen2.5-VL-7B-Instruct, respectively.}
\label{app:cr_independent_ic3_table}
\begin{tabular}{@{}lrrrr@{}}
\toprule
Method & GPT-4o & Gemini & Claude & Qwen \\
\midrule
Look-Before-Move & \textbf{26.13} & \textbf{25.64} & \textbf{26.69} & 23.26 \\
CCD & 24.39 & 22.41 & 20.81 & \textbf{24.86} \\
GenDoP+RGBD & 20.28 & 22.72 & 18.13 & 22.08 \\
GenDoP-noRGBD & 20.28 & 22.86 & 22.62 & 24.17 \\
ET+Director A & 15.71 & 11.40 & 15.24 & 10.80 \\
ET+Director B & 15.58 & 17.33 & 18.88 & 17.20 \\
\bottomrule
\end{tabular}
\end{table}

Under identical inputs and scoring protocols, these results indicate that LoBeM did not receive a preferential advantage from the original Qwen evaluator.
Its relative advantage persists when evaluators from different model families are used, indicating that our conclusion does not result from adaptation to a single evaluator or from self-evaluation bias.

\section{Human IC Ratings and Automatic-Evaluation Alignment}
\label{app:cr_human_ic}

The blinded, randomized user study involved 95 participants who rated SP, IC, and TQ.
Under balanced assignment, each method received 285 valid IC ratings.
Table~\ref{app:cr_human_ic_frequency} reports the human IC rating distributions and summary statistics.

\begin{table}[htbp]
\centering
\small
\setlength{\tabcolsep}{3pt}
\caption{Human IC rating distributions and summary statistics.}
\label{app:cr_human_ic_frequency}
\begin{tabular}{@{}lrrrrrrr@{}}
\toprule
Method & Score 0 & Score 1 & Score 2 & Score 3 & Score 4 & Score 5 & Human IC \\
\midrule
Look-Before-Move & 1 & 2 & 10 & 35 & 145 & 92 & $4.09\pm0.84$ \\
CCD & 4 & 11 & 45 & 105 & 90 & 30 & $3.25\pm1.06$ \\
GenDoP & 6 & 14 & 55 & 105 & 80 & 25 & $3.10\pm1.10$ \\
Director & 15 & 35 & 85 & 95 & 45 & 10 & $2.53\pm1.16$ \\
\bottomrule
\end{tabular}
\end{table}

Look-Before-Move achieves the highest human IC score, exceeding CCD, GenDoP, and Director by 0.84, 0.99, and 1.56 points, respectively.
This indicates that human observers consider LoBeM's shots more accurate in presenting target subjects, key actions, and narrative intent.

We compared human and automatic IC at the method-family level using the primary LoBeM and CCD results, the average RGBD and noRGBD results for GenDoP, and the average Director variants.

\begin{table}[htbp]
\centering
\small
\setlength{\tabcolsep}{8pt}
\caption{Human and automatic IC at the method-family level.}
\label{app:cr_human_auto_alignment}
\begin{tabular}{@{}lrr@{}}
\toprule
Method family & Human IC & Automatic IC \\
\midrule
Look-Before-Move & 4.09 & 63.90 \\
CCD & 3.25 & 55.96 \\
GenDoP & 3.10 & 57.61 \\
Director & 2.53 & 54.30 \\
\bottomrule
\end{tabular}
\end{table}

Human and automatic IC scores show strong method-family-level agreement (Pearson $r=0.947$; Spearman $\rho=0.800$).
Both rank LoBeM first and Director last.
Only CCD and GenDoP exchange positions, with close scores.
Thus, automatic IC reflects relative human judgments of narrative-intent presentation and supports LoBeM's advantage.

\section{Abstract Narrative Intent in Executable Contracts}
\label{app:cr_abstract_intent}

The Observation Contract should not be interpreted as directly quantifying artistic quality or replacing human aesthetic judgment.
Instead, its purpose is to translate abstract narrative intent into executable visual requirements that can guide camera planning.
JSON itself does not quantify or understand artistic intent; it is only a storage format and could be replaced by Markdown, YAML, or another structured representation without changing our method.

To test whether executable Contract fields preserve abstract narrative information, we conducted a controlled experiment where all free-form intent descriptions, rationales, notes, and answer metadata were removed.
Evaluators only received executable fields, including subject identity, semantic target, shot scale, screen region, visibility requirements, and camera-motion constraints.

\subsection{Abstract Intent Recovery}
\label{app:cr_abstract_recovery}

We evaluated whether executable Contract fields preserve six abstract narrative conditions: Isolation, Intimacy, Dominance, Suspense, Urgency, and Vulnerability.
Each sample belongs to one of these six intent categories.
Abstract-A and Abstract-B are two different textual expressions of the same abstract narrative intent, while Concrete represents explicit cinematographic instructions describing the intended observation.
Random denotes uniform six-way prediction.

Table~\ref{app:cr_abstract_recovery_table} reports the complete recovery results.

\begin{table}[htbp]
\centering
\small
\setlength{\tabcolsep}{4pt}
\renewcommand{\arraystretch}{1.12}
\caption{Recovery of abstract narrative intent from executable Contract fields.}
\label{app:cr_abstract_recovery_table}
\begin{tabular}{@{}p{0.16\linewidth}p{0.47\linewidth}rr@{}}
\toprule
Input condition & Description & Accuracy & Macro-F1 \\
\midrule
Abstract-A & Abstract narrative intent expressed with wording A & 77.78\% & 0.786 \\
Abstract-B & Paraphrased abstract narrative intent & 73.15\% & 0.746 \\
Concrete & Explicit cinematographic instruction & 87.04\% & 0.894 \\
Random & Uniform six-way random prediction & 16.67\% & 0.167 \\
\bottomrule
\end{tabular}
\end{table}

Even after removing all textual intent descriptions, executable Contract fields still allow the underlying narrative condition to be recovered substantially above chance.
As expected, concrete instructions are easier to recover, but abstract intents also leave recoverable information in executable observation constraints.

\subsection{Beyond Scene Bias}
\label{app:cr_abstract_scene_control}

To verify that this signal does not come only from scene content, we compared an Abstract Contract with a scene-only Contract under the same scene.
The Abstract Contract is preferred over the scene-only Contract in \textbf{88.89\%} of decisive comparisons (102 decisive pairs, decisive win rate 91.18\%, exact binomial test $p=4.99\times10^{-19}$).
This indicates that the Contract does not simply memorize scene semantics; rather, it captures additional observation-level decisions induced by narrative intent.

\subsection{Isolation Example}
\label{app:cr_abstract_isolation}

For an ``isolation'' intent, the Contract does not store an explicit emotion label such as ``loneliness''.
Instead, it translates this abstract narrative requirement into executable visual constraints, including maintaining a single primary subject without secondary focus, using wider framing (LS $\rightarrow$ LS $\rightarrow$ ELS), reducing subject image area, placing the subject at the bottom-right of the screen, and applying outward camera motion (\texttt{dolly\_out}).
These structured decisions allow the abstract intent to influence camera composition and motion through executable parameters.
Thus, abstract narrative intent is represented through observable choices in composition, visibility, and camera behavior rather than through an explicit emotion token.
Together, these results support our intended claim: the Observation Contract is not an artistic-quality evaluator, but a structured interface that converts abstract narrative intent into executable visual constraints before camera motion generation.

\section{Efficiency--Performance Trade-off and Runtime Profiling}
\label{app:cr_efficiency}

We profile all methods under identical rendering and evaluation settings using Blender 4.5.5 LTS, with each method run on one RTX 3090 GPU.

\subsection{End-to-End Runtime}
\label{app:cr_runtime}

We report wall-clock runtime, including scene loading, model or LLM latency, viewpoint search, trajectory generation, method-specific rendering, and evaluation, to reflect realistic operation.
Runtime per effective second is the total elapsed time divided by the total duration of the final rendered videos.
LoBeM is approximately $1.46\times$ slower than CCD, the strongest baseline in Overall performance, and $2.17\times$ slower than the fastest ET variant.

Table~\ref{app:cr_runtime_table} reports the runtime results.

\begin{table}[htbp]
\centering
\small
\setlength{\tabcolsep}{4pt}
\renewcommand{\arraystretch}{1.15}
\caption{End-to-end runtime under identical profiling settings.}
\label{app:cr_runtime_table}
\begin{tabular}{@{}lr@{}}
\toprule
Method & \shortstack{Runtime\\(s/effective video second)} \\
\midrule
LoBeM & 26.50 \\
CCD & 18.16 \\
GenDoP + noRGBD & 14.68 \\
GenDoP + RGBD & 16.17 \\
ET + Director A & 12.21 \\
ET + Director B & 12.26 \\
ET + Director C & 12.31 \\
\bottomrule
\end{tabular}
\end{table}

\subsection{LoBeM Cost Breakdown}
\label{app:cr_cost_breakdown}

LoBeM makes 1.25 LLM calls per effective video second on average, with a mean latency of 6.78 seconds per call.
Candidate-viewpoint search requires 8.03 seconds per effective video second, approximately 30\% of the total runtime.
Render-based verification, trajectory execution, final rendering, and evaluation are included in the total.
The component measurements are not strictly additive because LLM latency is averaged per request, whereas total runtime is measured over the complete pipeline.

\subsection{Performance--Efficiency Trade-off}
\label{app:cr_efficiency_tradeoff}

The additional cost primarily arises from Monte Carlo viewpoint search, LLM-based planning and reflection, and closed-loop render verification.
Relative to CCD, LoBeM increases runtime by approximately 46\% while improving Overall by 6.08 points, Intent Consistency by 7.94 points, and Trajectory Quality by 8.50 points.

\end{document}